\documentclass[sigconf]{acmart}

\usepackage{algorithm}
\usepackage{algpseudocode}
\usepackage{subcaption}
\usepackage{multirow}
\usepackage{graphicx}
\usepackage{threeparttable}
\usepackage{marvosym}

\AtBeginDocument{%
  \providecommand\BibTeX{{%
    \normalfont B\kern-0.5em{\scshape i\kern-0.25em b}\kern-0.8em\TeX}}}

\setcopyright{acmlicensed}
\copyrightyear{2024}
\acmYear{2024}
\acmDOI{XXXXXXX.XXXXXXX}
\renewcommand\footnotetextcopyrightpermission[1]{} 
\begin{document}

\title{LAPTOP-Diff: Layer Pruning and Normalized Distillation for Compressing Diffusion Models}

%
\author{Dingkun Zhang}
\authornote{Equal contribution.}
\authornote{Work done during his internship at OPPO.}
\affiliation{%
  \institution{Harbin Institute of\\ Technology, Shenzhen}
  \city{Shenzhen}
  \country{China}}
\email{dingkunzhang0xffff@gmail.com}

\author{Sijia Li}
\authornotemark[1]
\affiliation{%
  \institution{OPPO AI Center}
  \city{Shenzhen}
  \country{China}}
\email{lisijia@oppo.com}

\author{Chen Chen}
\authornote{Corresponding author.}
\affiliation{%
  \institution{OPPO AI Center}
  \city{Shenzhen}
  \country{China}}
\email{chenchen4@oppo.com}

\author{Qingsong Xie}
\affiliation{%
  \institution{OPPO AI Center}
  \city{Shenzhen}
  \country{China}}
\email{xieqingsong1@oppo.com}

\author{Haonan Lu}
\authornotemark[3]
\affiliation{%
  \institution{OPPO AI Center}
  \city{Shenzhen}
  \country{China}}
\email{luhaonan@oppo.com}

\renewcommand{\shortauthors}{Dingkun Zhang, et al.}



\begin{abstract}
In the era of AIGC, the demand for low-budget or even on-device applications of diffusion models emerged.
In terms of compressing the Stable Diffusion models (SDMs), several approaches have been proposed, and most of them leveraged the handcrafted layer removal methods to obtain smaller U-Nets, along with knowledge distillation to recover the network performance.
However, such a handcrafting manner of layer removal is inefficient and lacks scalability and generalization, and the feature distillation employed in the retraining phase faces an imbalance issue that a few numerically significant feature loss terms dominate over others throughout the retraining process.
To this end, we proposed the layer pruning and normalized distillation for compressing diffusion models (LAPTOP-Diff). We, 1) introduced the layer pruning method to compress SDM's U-Net automatically and proposed an effective one-shot pruning criterion whose one-shot performance is guaranteed by its good additivity property, surpassing other layer pruning and handcrafted layer removal methods, 2) proposed the normalized feature distillation for retraining, alleviated the imbalance issue.
Using the proposed LAPTOP-Diff, we compressed the U-Nets of SDXL and SDM-v1.5 for the most advanced performance, achieving a minimal 4.0\% decline in PickScore at a pruning ratio of 50\% while the comparative methods' minimal PickScore decline is 8.2\%.
\end{abstract}




\begin{teaserfigure}
    \centering
    Images generated with the original 2.6B U-Net\\
    \includegraphics[width=\textwidth]{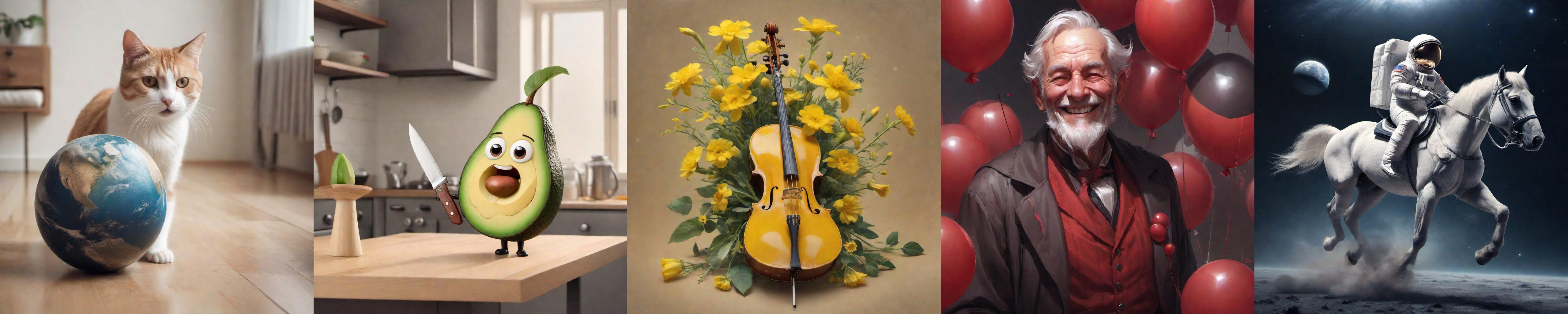}\\
    Images generated with our pruned 1.3B U-Net\\
    \includegraphics[width=\textwidth]{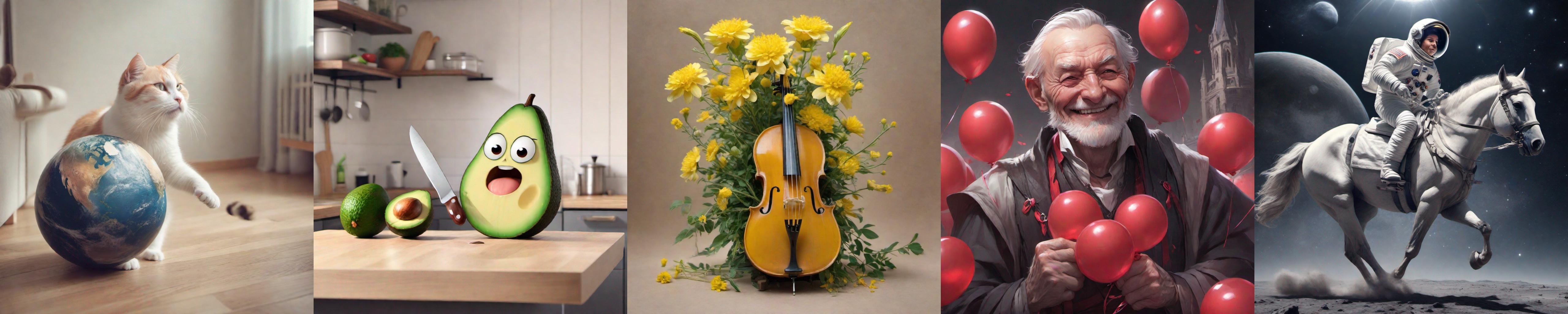}\\
    \caption{Visual comparison between the original model and our 50\% pruned model. Our 1.3B U-Net is pruned and retrained from ZavychromaXL-v1.0.}
    \label{fig_teaser}
\end{teaserfigure}


\maketitle

\section{Introduction}
Generative modeling for text-to-image (T2I) synthesis has experienced rapid progress in recent years.
Especially, the diffusion model\cite{ddpm, sde, dm_beats_gan, ldm, sdxl} emerged with its capability of generating high-resolution, photorealistic, and diverse images.
Among all the diffusion models, Stable Diffusion model (SDM)\cite{ldm, sdxl} is the most influential one, playing a pivotal role in the AIGC community as an open-source framework, serving as the foundation of a wide spectrum of downstream applications.

However, SDM's remarkable performance comes with its considerable memory consumption and latency, making its deployment on personal computers or even mobile devices highly constrained.
Furthermore, recent versions of the SDM series, e.g., SDXL\cite{sdxl}, tend to grow larger in parameters, resulting in even greater memory consumption and latency.
To reduce the SDM's inference budget, several approaches have been proposed, i.e., denoising step reduction\cite{pd, lcm, rectified_flow, ufogen, add}, efficient architecture designing\cite{mobile_diffusion, bk_sdm, ssd, koala}, structural pruning\cite{structural_pruning_diff}, quantization\cite{ptq_diff, q_diff}, and hardware-aware optimization\cite{speed_is_all_you_need}.
These approaches are generally orthogonal to each other.

Among these approaches, efficient architecture designing and structural pruning are understudied.
On the one hand, the previous efficient architecture designing approaches\cite{mobile_diffusion, bk_sdm, ssd, koala} typically go through substantial empirical studies to identify the unimportant layers of SDM's U-Net and then remove them to achieve a smaller and faster network.
Such a handcrafting manner usually can not achieve the best performance and lacks scalability and generalization.
We noticed that these hand-crafted layer-removal methods can be substituted by the layer-pruning method in an automatic scheme for better scalability and performance.
On the other hand, the previous structural pruning approach\cite{structural_pruning_diff} on SDM focused on the fine-grained pruning, i.e., pruning rows and columns of parameter matrices.
However, there are studies\cite{filter_prune_or_layer_prune, shortened_llama} suggest that the fine-grained pruning is generally less efficient in reducing model latency compared to the coarser-grained layer pruning, and interestingly, layer pruning is possible to achieve the same performance as finer-grained structural pruning or even better.
Based on the above two points, the layer pruning method is worth studying.

After the layer removal or pruning, the SDMs often cannot generate clear images directly.
Previous methods\cite{mobile_diffusion, bk_sdm, ssd, koala} exploit knowledge distillation to retrain the pruned network to recover its performance.
Previous methods typically leveraged three types of objectives, i.e., regular training objective, logit distillation(output distillation)\cite{logit_distill} objective, and feature distillation\cite{feature_distill} objective.
Among the three parts, feature distillation is the pivotal one.
However, our further examinations revealed an imbalance issue in the previous distillation-based retraining method that a few feature loss terms dominated over others throughout the retraining process, leading to degraded performance.

In this work, we proposed the LAyer Pruning and normalized disTillation for cOmPressing Diffusion models(LAPTOP-Diff), pushing the realms of efficient architecture designing and structural pruning for SDM toward automation, scalability, and greater performance.
We formulated the problem of layer pruning from a high perspective of the combinatorial optimization problem and solved it in a simple yet effective one-shot manner.
Benefiting from such perspective, we explored several other possible pruning criteria, and through ablation studies, we were able to discover that the effectiveness of our one-shot layer pruning criterion comes from its good additivity property\cite{additivity_1, additivity_2}.
Further, we identified an imbalance issue in the previous distillation-based retraining method that a few feature loss terms dominated over others throughout the retraining process, and we alleviated this issue through the proposed normalized feature distillation.
Our contributions are summarized as follows:
\begin{itemize}
    \item We explored layer pruning, an understudied approach of structural pruning, on SDMs, and proposed an effective one-shot pruning criterion whose one-shot performance is guaranteed by its good additivity property, surpassing other layer pruning and handcrafted layer removal methods, pushing the previous layer-remove-based efficient architecture designing approaches toward automation, scalability, and greater performance.
    \item We alleviated the imbalance issue of the previous distillation-based retraining using the proposed normalized feature distillation.
    \item Our proposed LAPTOP-Diff greatly surpasses the layer-remove-based efficient architecture designing approaches in terms of network performance across different SDMs and pruning ratios.
\end{itemize}

\section{Related Work}
\subsection{Diffusion Model}
Diffusion model\cite{ddpm, sde, dm_beats_gan, ldm, sdxl} is a type of generative model that leverages iterative denoising process to synthesize data.
In the realm of text-to-image(T2I) synthesis, the diffusion models such as DALL·E\cite{dalle}, Imagen\cite{imagen}, Deepfloyd IF\cite{deepfloyd}, and Stable Diffusion\cite{ldm} demonstrated their remarkable capability of generating high-resolution, photo-realistic, and diverse images.
Among various diffusion models, Stable Diffusion\cite{ldm} is the most influential one in both the academic community and the industry.
Stable Diffusion model(SDM) is a type of Latent Diffusion model that performs the iterative denoising process in the low-dimensional latent space and then transforms the latent representations into pixel-space images through a VAE decoder.
There is also a more recent version of the SDM series, i.e., SDXL\cite{sdxl}, which demonstrates superior image generation quality at a higher resolution of 1024$\times$1024.

However, the impressive performance of diffusion models comes with substantial memory consumption and latency.
To reduce the model budget of SDM, several approaches have been explored, e.g., denoising step reduction\cite{pd, lcm, rectified_flow, ufogen, add}, quantization\cite{ptq_diff, q_diff}, hardware-aware optimization\cite{speed_is_all_you_need}, efficient architecture designing\cite{mobile_diffusion, bk_sdm, ssd, koala}, and structural pruning\cite{structural_pruning_diff}.

\subsection{Efficient Architecture Designing for SDM}
Orthogonal to many other approaches of reducing SDM's model budget, efficient architecture designing mainly aims to design an efficient sub-structure of the original SDM's U-Net since most of the SDM's memory consumption and latency comes from its U-Net.
Previous methods\cite{mobile_diffusion, bk_sdm, ssd, koala} of this kind typically go through substantial empirical studies to identify the unimportant layers of the SDM's U-Net and remove them to achieve a smaller and faster network.
BK-SDM\cite{bk_sdm} handcrafted 3 efficient U-Nets of different sizes for SDM-v1 or SDM-v2 by layer removal, partially following the empirical conclusion of DistilBERT\cite{distilbert} which is an empirical work on compressing the BERT model.
SSD-1B\cite{ssd} and Segmind-Vega\cite{ssd} (we refer to them as SSD and Vega for the rest of this paper) handcrafted 2 efficient U-Nets of different sizes for SDXL by identifying the unimportant layers through human evaluation and then imposing layer removal.
KOALA\cite{koala} is derived from BK-SDM and handcrafted 2 efficient U-Nets of different sizes for SDXL also by layer removal.

Such a handcrafting manner usually can not achieve the best performance and lacks scalability and generalization.
We noticed that these layer-remove-based methods can be categorized as handcrafted layer pruning.
Hence, we propose that these handcrafted layer-remove methods can be substituted by the layer pruning method in an automatic scheme for better scalability and performance.

\subsection{Layer Pruning}
Layer pruning\cite{layer_prune_based_on_feature_representations, filter_prune_or_layer_prune, shortened_llama}, also known as depth pruning, is a type of structural pruning method that aims to automatically evaluate and remove unimportant layers.
Unlike other structural pruning methods, layer pruning is paid little attention due to its coarse-grained nature.
Layer pruning is often considered less effective compared to those fine-grained structural pruning methods.
However, there are studies\cite{filter_prune_or_layer_prune, shortened_llama} showed that, compared to fine-grained structural pruning methods, coarse-grained layer pruning is generally more efficient in reducing model's latency and possible to achieve the same performance or even better.

Although there are several layer pruning methods\cite{layer_prune_based_on_feature_representations, filter_prune_or_layer_prune, shortened_llama} have been proposed, none of which viewed layer pruning through a combinatorial optimization problem perspective.
Magnitude-based and Taylor-expansion-based pruning are common baselines.
Magnitude-based layer pruning uses the summed magnitude of the parameters in a layer as the layer importance criterion, and Taylor-expansion-based layer pruning uses the first-order Taylor-expansion of the loss function as the layer importance criterion.
After evaluating the importance of each layer by different importance criteria, the previous layer pruning methods then opt to prune the least important layers.

In this work, we formulated the problem of layer pruning from a higher perspective of the combinatorial optimization problem and solved it in a simple yet effective one-shot manner, forming an effective one-shot pruning criterion, surpassing other layer pruning and handcrafted layer removal methods.
Furthermore, through such a perspective, we were able to identify that the effectiveness of our one-shot layer pruning criterion comes from its good additivity property\cite{additivity_1, additivity_2}.

\subsection{Distillation-Based Retraining}
After the layer removal or pruning, the SDMs often can not generate clear images directly.
Previous methods\cite{mobile_diffusion, bk_sdm, ssd, koala} retrain the pruned SDMs by utilizing knowledge distillation to recover their performance.
Knowledge distillation used in the retraining phase usually consists of three parts, i.e., regular training objective, logit distillation(output distillation)\cite{logit_distill}, and feature distillation\cite{feature_distill}.
Among the three parts, feature distillation is the most effective one.

However, in practice, we located an imbalance issue in the distillation-based retraining process.
To this end, we proposed a simple yet effective re-weighting strategy to alleviate this issue.

\begin{figure*}[htp]
	\centering
    \includegraphics[width=\textwidth]{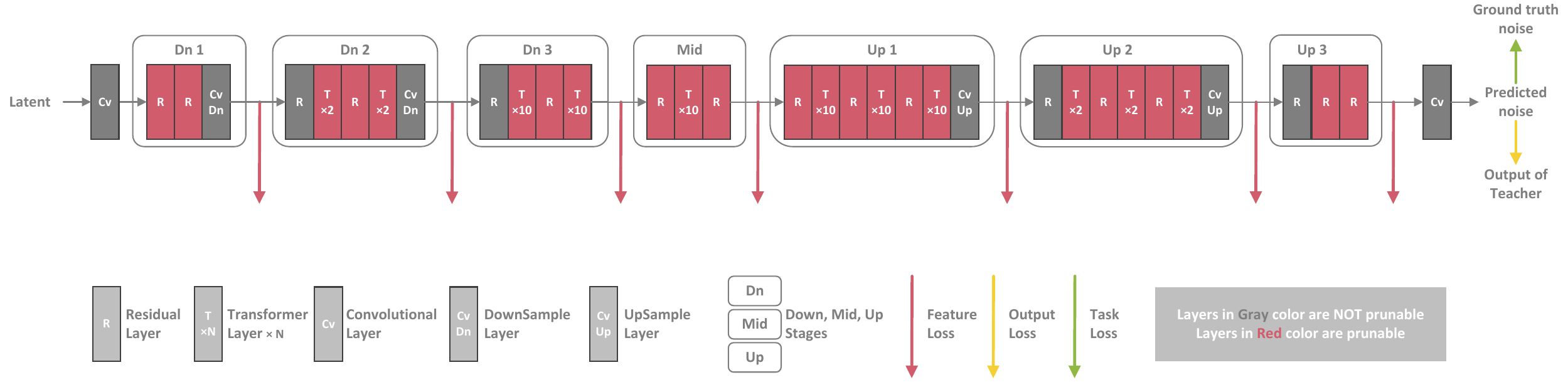}
	\caption{Architecture of SDM's U-Net and knowledge distillation framework, using SDXL as an example. Other SDMs' U-Net have slightly different architectures, e.g., SDM-v1.5 has 4 Dn and Up stages instead of 3 and has fewer transformer layers.}
\label{fig_UNet}
\end{figure*}

\section{Methodology}
\subsection{Preliminary}
The SDM's U-Net can be viewed as a stack of residual layers, transformer layers, and several up/down sampling convolutional layers, omitting the skip connections.
The majority of these layers are consistent in the shape of their own inputs and outputs.
Thus, these layers can be removed directly without causing a shape mismatch problem.
Specifically, except for several residual layers that involve alterations of the number of input and output channels, all the residual layers and transformer layers of SDM's U-Net are prunable as shown in Fig. \ref{fig_UNet}.
The prunable layers account for 88\% and 98\% total parameters of the entire U-Net of SDM-v1.5 and SDXL respectively.
We denote the collection of all the $n$ prunable layers as $L^n=\{l_1, \dots, l_n\}$ where $l_i$ is a single residual layer or transformer layer.

The previous approaches\cite{bk_sdm, ssd} of retraining the pruned SDM's U-Net leverage three objectives, i.e., task loss, output distillation\cite{logit_distill} loss and feature distillation\cite{feature_distill} loss:
\begin{equation}
\begin{aligned}
\mathcal{L}_{Task} &= \mathbb{E}_{z, y, t, \epsilon}\left|\left|\epsilon-\epsilon_S(z_t, y, t)\right|\right|_2^2 , \\
\mathcal{L}_{OutKD} &= \mathbb{E}_{z, y, t}\left|\left|\epsilon_T(z_t, y, t)-\epsilon_S(z_t, y, t)\right|\right|_2^2 , \\
\mathcal{L}_{FeatKD} &= \sum_i\mathbb{E}_{z, y, t}\left|\left|f^i_T(z_t, y, t)-f^i_S(z_t, y, t)\right|\right|_2^2  ,
\end{aligned}
\end{equation}
where $z$, $y$ and $t$\textasciitilde$Uniform(0, T)$ denote the latent representation, its corresponding condition, and its corresponding time step. $\epsilon$\textasciitilde$N(0, I)$ denote the ground-truth noise.
$\epsilon_S$ and $\epsilon_T$ are the outputs of the student and the teacher where the student is the pruned U-Net and the teacher is the frozen original U-Net.
$f^i_T$ and $f^i_S$ represent the feature maps at the end of the $i$-th stage in the teacher and the student, respectively.
We refer to the $\mathbb{E}_{z, y, t}\left|\left|f^i_T(z_t, y, t)-f^i_S(z_t, y, t)\right|\right|_2^2$ terms for different $i$ as feature loss terms in this paper.
The $\mathcal{L}_{OutKD}$ and $\mathcal{L}_{FeatKD}$ are knowledge distillation objectives.
The three objectives are also illustrated in Fig. \ref{fig_UNet}.
The overall objective is $\mathcal{L}_{all} = \mathcal{L}_{Task} + \lambda_{OutKD}\mathcal{L}_{OutKD} + \lambda_{FeatKD}\mathcal{L}_{FeatKD}$.
Where the $\lambda_{OutKD}$ and $\lambda_{FeatKD}$ are hyperparameters, both of which are usually set to 1 as default.

\subsection{One-Shot Layer Pruning}
The goal of layer pruning is to remove a subset of prunable layers from the U-Net that satisfies a given pruning ratio while causing a minimal decrease in network performance.
We opt to minimize the Mean-Squared-Error loss between the final outputs of the original network and the pruned network given a pruning ratio.
There are also other possible objectives, and we will discuss them at the end of this section.
Let $\epsilon_{ori}$ denote the output of the original network, $r_i \in L^n$ denote the $i-th$ removed layer, and $\epsilon(r_1, r_2, \dots, r_m)$ denote the output of the network where the layers $r_1, r_2, \dots, r_m$ are removed.
Notice that the $m$ is an undetermined variable.
The optimization problem is formulated as follows:
\begin{equation}
\begin{aligned}
\mathop{\min}_{r_1, \ldots, r_m}
\mathbb{E}||&\epsilon(r_1, \ldots, r_m)-\epsilon_{ori}||_2^2 \\
s.t.\ \{r_1, \ldots, r_m\} &\subset L^n,\ \sum_{i=1}^{m}params(r_i) \geq P,
\end{aligned}
\label{objective_1}
\end{equation}
where the $param(r_i)$ denotes the total count of parameters of layer $r_i$, and $P$ is the number of parameters to prune which is calculated by the total number of parameters of the original network multiplied by the pruning ratio.
Directly solving this optimization problem requires the search on all combinations of $\{r_1, r_2, \dots, r_m\}$ which is NP-hard.
Therefore, we try to find a surrogate objective.

\textbf{Upper bound of objective (\ref{objective_1}).}
According to the triangle inequality, we have an upper bound of formula (\ref{objective_1}):
\begin{equation}
\begin{aligned}
\mathbb{E}&\left|\left|\epsilon(r_1, \ldots, r_m)-\epsilon_{ori}\right|\right|_2^2 \\
\leq&\mathbb{E}\left|\left|\epsilon(r_1)-\epsilon_{ori}\right|\right|_2^2 \\
&+\mathbb{E}\left|\left|\epsilon(r_1, r_2)-\epsilon(r_1)\right|\right|_2^2 \\
&+\ldots \\
&+\mathbb{E}\left|\left|\epsilon(r_1, \ldots, r_m)-\epsilon(r_1, \ldots, r_{m-1})\right|\right|_2^2.
\end{aligned}
\label{inequality_1}
\end{equation}
Therefore, we find a surrogate objective for the objective (\ref{objective_1}):
\begin{equation}
\begin{aligned}
\mathop{\min}_{r_1, \ldots, r_m}
&\mathbb{E}\left|\left|\epsilon(r_1)-\epsilon_{ori}\right|\right|_2^2 +\mathbb{E}\left|\left|\epsilon(r_1, r_2)-\epsilon(r_1)\right|\right|_2^2+\ldots \\
&+\mathbb{E}\left|\left|\epsilon(r_1, \ldots, r_m)-\epsilon(r_1, \ldots, r_{m-1})\right|\right|_2^2 \\
s.t.\ \{r_1, &\ldots, r_m\} \subset L^n,\ \sum_{i=1}^{m}params(r_i) \geq P.
\end{aligned}
\label{objective_2}
\end{equation}
Solving the optimization problem (\ref{objective_2}) is still NP-hard. However, it led us to another, better surrogate objective.

\textbf{Approximation of upper bound (\ref{objective_2}).}
We noticed that each term of the formula (\ref{objective_2}) is the Mean-Squared-Error loss between a network(pruned or unpruned) and the network where one additional layer is removed,
e.g, $\left|\left|\epsilon(r_1, \ldots, r_{i-1}, r_i)-\epsilon(r_1, \ldots, r_{i-1})\right|\right|_2^2$ is the Mean-Squared-Error loss between the network where layers $r_1, \ldots, r_{i-1}$ are removed and the network where layers $r_1, \ldots, r_{i-1}, r_i$ are removed.
We conjecture that the Mean-Squared-Error loss caused by removing the same layer $r_i$ are similar across different networks.
The assumption is formulated as follows:
\begin{equation}
\begin{aligned}
&\mathbb{E}\left|\left|\epsilon(r_1, \ldots, r_{i-1}, r_i)-\epsilon(r_1, \ldots, r_{i-1})\right|\right|_2^2 \\
&\approx\mathbb{E}\left|\left|\epsilon(r_1, \ldots, r_{i-2}, r_i)-\epsilon(r_1, \ldots, r_{i-2})\right|\right|_2^2 \\
&\approx\dots \\
&\approx\mathbb{E}\left|\left|\epsilon(r_1, r_i)-\epsilon(r_1)\right|\right|_2^2 \\
&\approx\mathbb{E}\left|\left|\epsilon(r_i)-\epsilon_{ori}\right|\right|_2^2.
\end{aligned}
\label{approximation_1}
\end{equation}
Finally with assumption (\ref{approximation_1}), we can optimize the objective (\ref{objective_2}) by optimizing its approximated objective:
\begin{equation}
\begin{aligned}
\mathop{\min}_{r_1, \ldots, r_m}
&\sum_{i=1}^{m}\mathbb{E}\left|\left|\epsilon(r_i)-\epsilon_{ori}\right|\right|_2^2 \\
s.t.\ \{r_1, \ldots, r_m\} &\subset L^n,\ \sum_{i=1}^{m}params(r_i) \geq P.
\end{aligned}
\label{objective_3}
\end{equation}
Using the objective (\ref{objective_3}) to surrogate the original objective (\ref{objective_1}) implies an interesting property called additivity\cite{additivity_1, additivity_2} that a network's output distortion caused by many perturbations approximately equals to the sum of output distortion caused by each single perturbation.
Further experiments in Sec~\ref{sec_ablation_additivity} validated the additivity property of our method on SDMs, showing that the final objective (\ref{objective_3}) is an excellent approximation of the original objective (\ref{objective_1}), meanwhile, the assumption (\ref{approximation_1}) is well supported.

\begin{algorithm}[t]
\caption{One-Shot Layer Pruning}
\label{algo_one_shot_pruning}
\begin{algorithmic}[1]
\renewcommand{\algorithmicrequire}{\textbf{Input:}}
\renewcommand{\algorithmicensure}{\textbf{Output:}}
\Require{A network with prunable layers $L^n = \{l_0, l_1, \dots, l_n\}$, a calibration dataset X, parameters of total quantity $P$ to prune}
\Ensure{A set of layers to prune $R^m \subset L^n$}
\For{$l_i$ in $L^n$}
    \State calculate $values[i] = \mathbb{E}\left|\left|\epsilon(l_i)-\epsilon_{ori}\right|\right|_2^2$ on dataset X
    \State $weights[i] = params(l_i)$
\EndFor
\State $knapsack\_capacity = P$
\State acquires $R^m$ by solving the variant 0-1 Knapsack problem (\ref{objective_3}) given $values, weights, knapsack\_capacity$, using dynamic programming\\
\Return $R^m$
\end{algorithmic}
\end{algorithm}

Here, the term $\mathbb{E}\left|\left|\epsilon(r_i)-\epsilon_{ori}\right|\right|_2^2$ functions as a pruning criterion.
We only need to compute $\mathbb{E}\left|\left|\epsilon(l_i)-\epsilon_{ori}\right|\right|_2^2$, i.e., the output loss between the original network and the network where only layer $l_i$ is removed, for each layer $l_i \in L^n$ with the total time complexity of $O(n)$,
then the optimization problem (\ref{objective_3}) becomes a variant of the 0-1 Knapsack problem.
Similar to the original 0-1 Knapsack problem, this variant problem can also be solved by the dynamic programming algorithm or can be approximately solved by the greedy algorithm.
Both pseudocodes are given in the supplementary.
When using the greedy algorithm, the solution degenerates to recurrently pruning the least important layer until the condition $\sum_{i=1}^{m}params(r_i) \geq P$ is satisfied, the same scheme as the previous approaches adopted.
The greedy algorithm is simpler to implement, but the dynamic programming algorithm is always better.
We used the dynamic programming algorithm in our work.
Finally, the one-shot layer pruning method is summarized in Algorithm.\ref{algo_one_shot_pruning}.

\textbf{Discussions.}
In this section, we formulated the one-shot layer pruning problem as a combinatorial optimization problem and derived a pruning criterion, i.e., the output loss $\mathbb{E}\left|\left|\epsilon(l_i)-\epsilon_{ori}\right|\right|_2^2$.
Furthermore, using the similar derivation form formula (\ref{objective_1}) to formula (\ref{objective_3}), we can construct some other possible pruning criteria, i.e., $\Delta$task loss, $\Delta$CLIP score\cite{clipscore}, calculated by $\mathbb{E}|\mathcal{L}(r_i)-\mathcal{L}_{ori}|$ and $\mathbb{E}|\mathcal{C}(r_i)-\mathcal{C}_{ori}|$ respectively, where the $\mathcal{L}$ is the task loss and the $\mathcal{C}$ is the CLIP score\cite{clipscore}.
Derivations of these criteria are in the supplementary.
Experiments in Sec~\ref{sec_ablation_additivity} and ablation studies in Sec~\ref{sec_ablation_criteria} showed that our output loss pruning criterion significantly satisfies the additivity property and achieves the best performance among all the pruning criteria on different SDM models.
Further, among the three pruning criteria we constructed, i.e., output loss, $\Delta$task loss, and $\Delta$CLIP score, for each model, the criteria with stronger additivity properties in Sec~\ref{sec_ablation_additivity} achieved better pruning performance in Sec~\ref{sec_ablation_criteria}.
This observation is justifiable since the one-shot layer pruning directly optimizes the surrogate objectives such as (\ref{objective_3}) rather than the original objective such as (\ref{objective_1}), optimizing those surrogate objectives that better approximate the original objectives can generally achieve better pruning performance.
Based on the above observation and discussion, we can conclude that the effectiveness of our output loss criterion comes from its good additivity property.

\begin{figure}[t]
	\centering
    \includegraphics[width=\columnwidth]{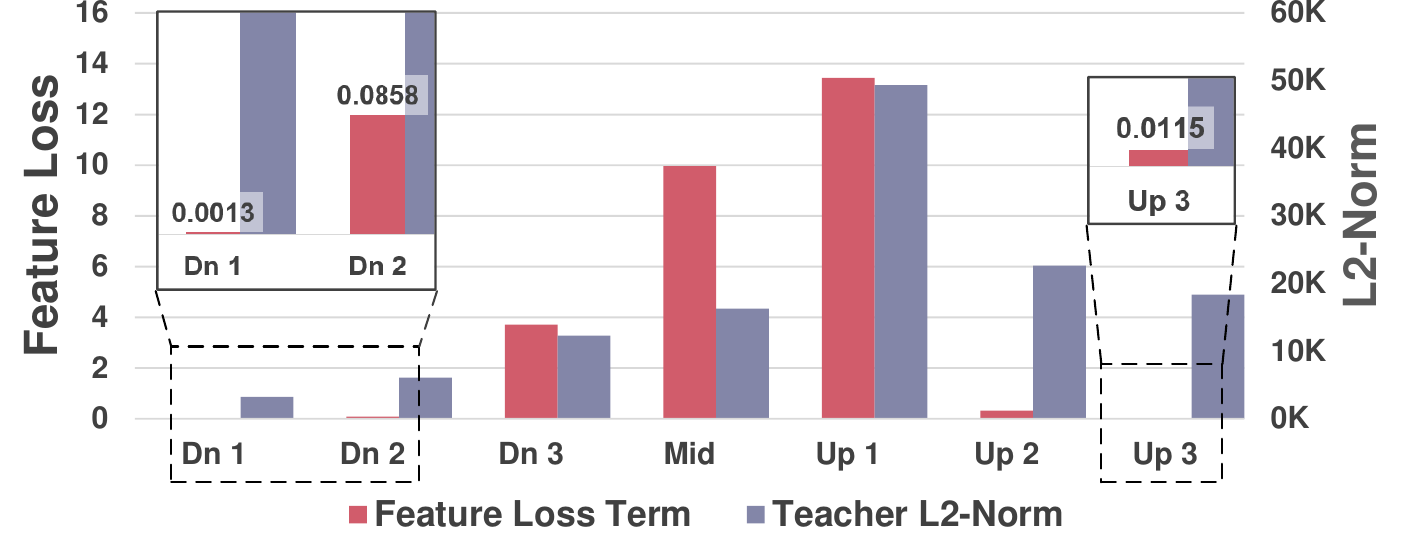}
	\caption{L2-Norms of feature maps at the end of each stage of the SDXL teacher and the feature loss terms at the 15K-th iteration of retraining.}
\label{fig_norm}
\end{figure}
\subsection{Normalized Feature Distillation}
In practice, we find that the pivotal objective for retraining is the feature loss $\mathcal{L}_{FeatKD}$.
However, our further examinations revealed an imbalance issue in the previous feature distillation approaches employed in the retraining phase.
As shown in Fig.\ref{fig_norm}, both the L2-Norms of feature maps at the end of different stages and the values of different feature loss terms vary significantly.
The same phenomenon shown in Fig.\ref{fig_norm} is observed on both SDXL and SDM-v1.5 under different pruning settings using either our layer pruning or handcrafted layer removal.
The highest feature loss term is around 10000 times greater than the lowest one throughout the distillation process, and produces around 1000 times larger gradients, diluting the gradients of the numerically insignificant feature loss terms.
Since the feature maps with greater L2-Norms naturally tend to produce greater feature loss terms, the significant magnitude difference of different feature loss terms is attributed to two factors, i.e., at the end of different stages, the inherent dissimilarity of the feature maps between teacher and student varies, and the L2-Norm of the feature maps varies.
The second factor aggravates the imbalance issue to a great extent.

Based on the above observations, simply adding up all the feature loss terms will cause the severe domination of a few feature loss terms over others, hindering the numerically insignificant feature loss terms from decreasing, leading to degraded performance.
To this end, we proposed a simple yet effective re-weighting strategy for feature distillation to eliminate the second factor's impact on the imbalance issue.
We opt to leverage the L2-Norms of the teacher's feature maps to re-weight the feature loss terms and adapt the feature distillation to our pruning scheme.
The normalized feature loss is formulated as:
\begin{equation}
\begin{aligned}
\mathcal{L}_{FeatKD-normed} &= \sum_{i\in V}\mathbb{E}_{z, y, t}~\alpha^i\left|\left|f^i_T(z_t, y, t)-f^i_S(z_t, y, t)\right|\right|_2^2\\
&\alpha^i = \frac{\sum_{j\in V}\left|\left|f^j_T(z_t, y, t)\right|\right|_2}{\left|V\right|\left|\left|f^i_T(z_t, y, t)\right|\right|_2},
\end{aligned}
\end{equation}
where $V$ is the set of stages where there are still residual layers or transformer layers remained after pruning, and $\left|V\right|$ is the size of set $V$.
Hence, our overall retraining objective is:
\begin{equation}
\begin{aligned}
\mathcal{L}_{all} = \mathcal{L}_{Task} + \lambda_{OutKD}\mathcal{L}_{OutKD} + \lambda_{FeatKD}\mathcal{L}_{FeatKD-normed}.
\end{aligned}
\end{equation}
Without any hyperparameter tuning, we set $\lambda_{OutKD}$ and $\lambda_{FeatKD}$ to 1.

\begin{figure}[t]
	\centering
	\subfloat[Our 50\% pruned SDXL.]{\includegraphics[width=0.49\columnwidth]{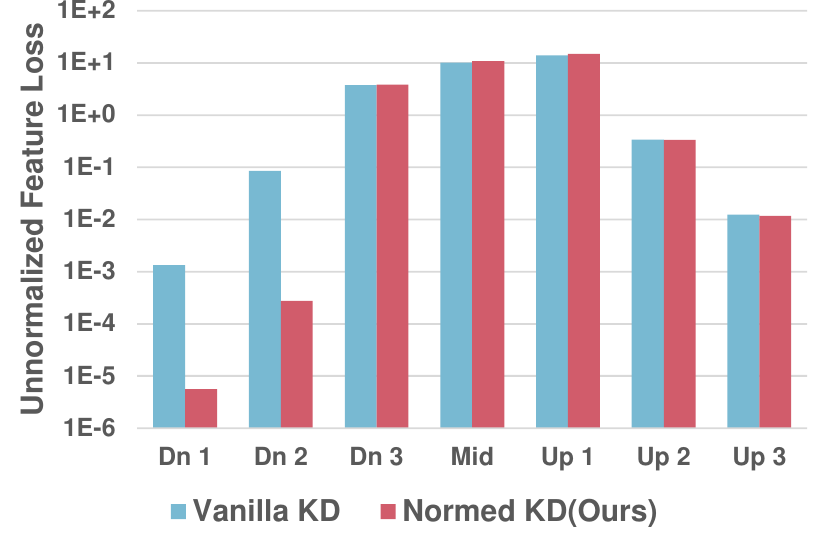}}\hspace{2pt}
	\subfloat[Our 33\% pruned SDM-v1.5]{\includegraphics[width=0.49\columnwidth]{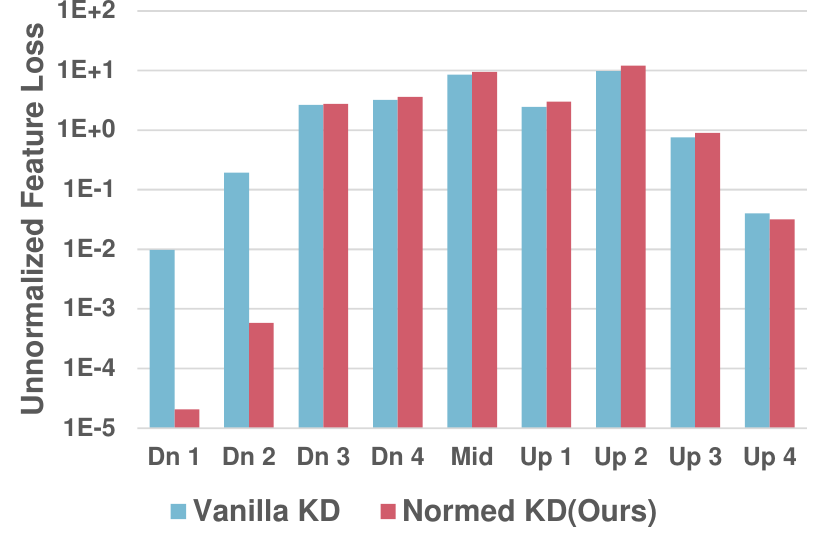}}
	\caption{Unnormalized feature loss terms, i.e., $\left|\left|f^i_T-f^i_S\right|\right|_2^2$, of each stage using vanilla feature distillation or our normalized feature distillation, at the 15K-th iteration of retraining.}
\label{fig_fealoss_ablation}
\end{figure}

\textbf{Discussions.}
We further investigated our method's impact on different $\left|\left|f^i_T-f^i_S\right|\right|_2^2$ terms during retraining.
As shown in Fig.\ref{fig_fealoss_ablation}, by applying the normalized feature distillation, we observed significant declines of the $\left|\left|f^i_T-f^i_S\right|\right|_2^2$ terms at some stages with marginal increasements elsewhere, proving that the dominance of a few feature loss terms over others is alleviated.
Further ablation studies in Sec~\ref{sec_ablation_distill} showed that our method finally achieved great performance improvements.

\begin{figure*}[t]
	\centering
    \begin{minipage}{0.1\textwidth}
    \quad
    \end{minipage}\hspace{10pt}
    \begin{minipage}{0.8\textwidth}
	\ \ ZavychromaXL-v1.0 \qquad\ SSD 1331M \qquad\qquad Ours 1316M \qquad\qquad Vega 745M \qquad\qquad\ Ours 729M
    \end{minipage}\\
    
    \begin{minipage}{0.1\textwidth}
    a cute Shiba Inu head in a cabbage
    \end{minipage}\hspace{10pt}
    \begin{minipage}{0.8\textwidth}
	\subfloat{\includegraphics[width=\textwidth]{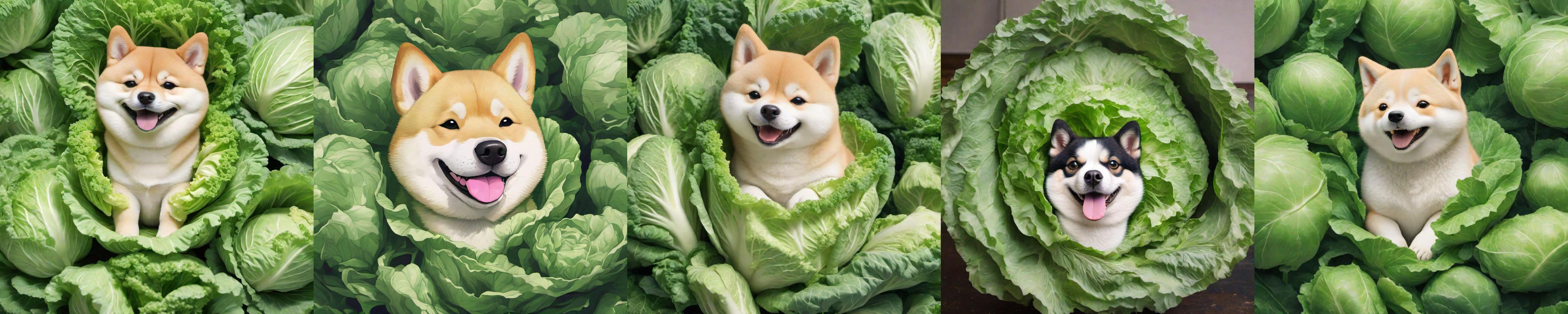}}
    \end{minipage}\\
    
    \begin{minipage}{0.1\textwidth}
    a giant cat carrying a small castle on its back, painting
    \end{minipage}\hspace{10pt}
    \begin{minipage}{0.8\textwidth}
	\subfloat{\includegraphics[width=\textwidth]{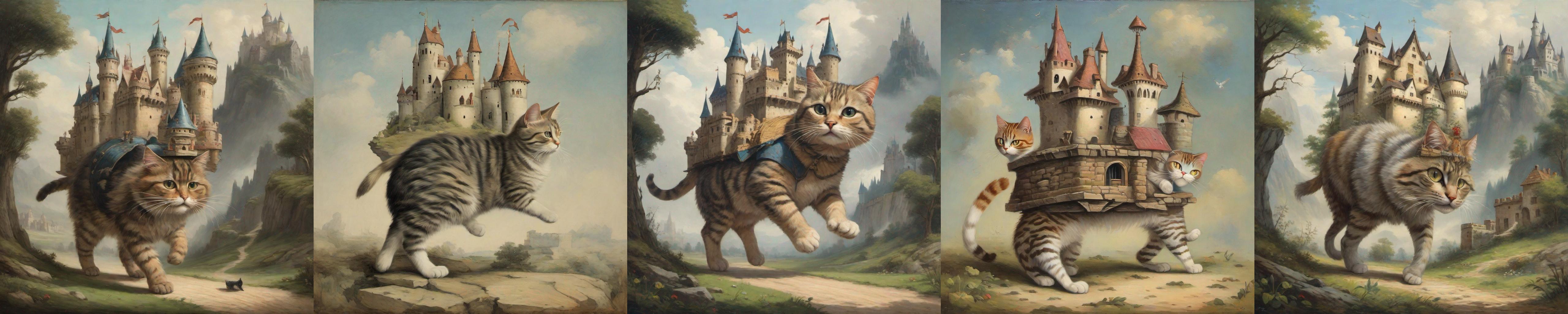}}
    \end{minipage}\\
    
    \begin{minipage}{0.1\textwidth}
    a classical guitar player at the concert
    \end{minipage}\hspace{10pt}
    \begin{minipage}{0.8\textwidth}
	\subfloat{\includegraphics[width=\textwidth]{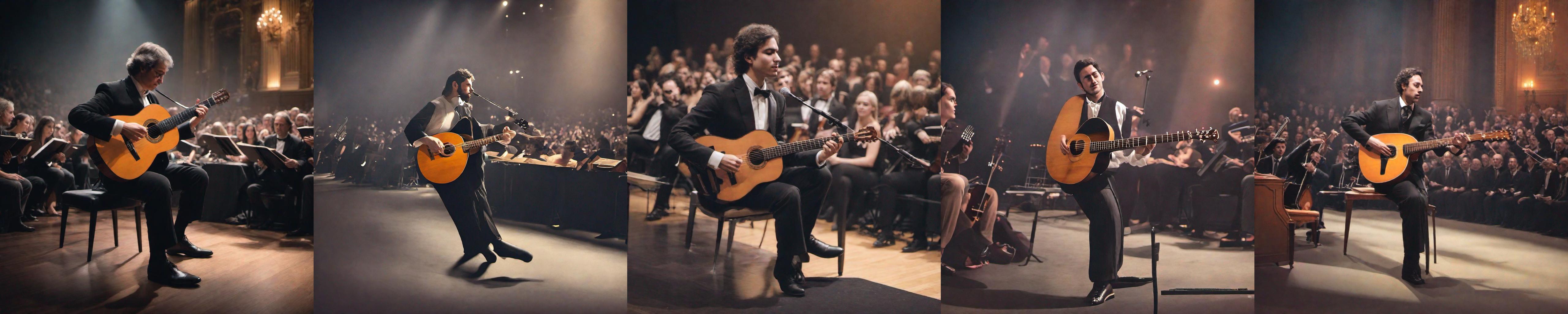}}
    \end{minipage}\\
    
    \begin{minipage}{0.1\textwidth}
    two tango dancers are dancing on the dim stage
    \end{minipage}\hspace{10pt}
    \begin{minipage}{0.8\textwidth}
	\subfloat{\includegraphics[width=\textwidth]{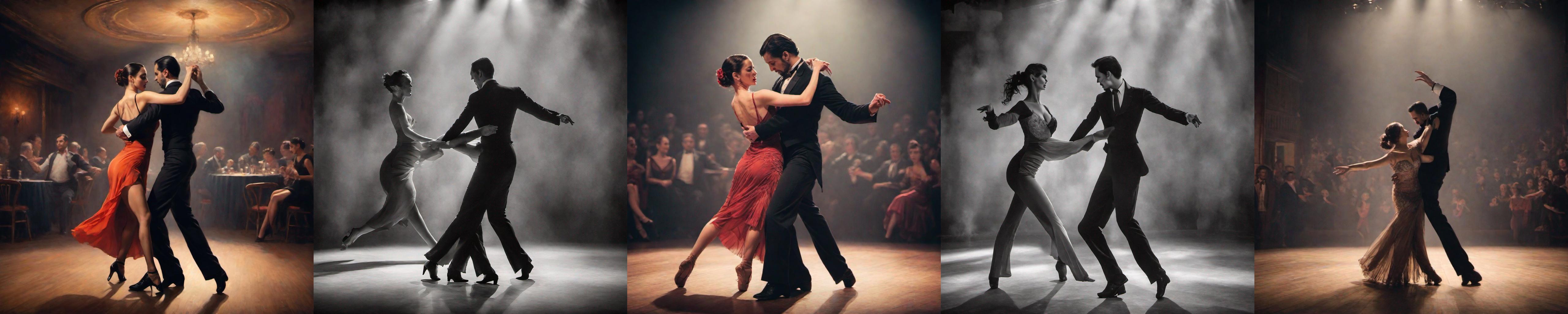}}
    \end{minipage}\\
\caption{Visual comparison with SSD and Vega in DDIM 25 steps.}
\label{fig_visual_comp}
\end{figure*}

\begin{table*}[t]
\centering
\scalebox{0.9}{
    \begin{tabular}{cccccc}
    \hline
    \multicolumn{2}{c}{Method} & Params of U-Net & HPS v2$\uparrow$ & PickScore$\uparrow$ & ImageReward$\uparrow$ \\ \hline
    \multirow{5}{*}{\begin{tabular}[c]{@{}c@{}}SDXL\\ Teacher: ZavychromaXL-v1.0\cite{zavy}\end{tabular}} & Teacher & 2567M & 0.27968 & 0.2755 & 0.9950 \\ \cline{2-6} 
     & SSD\cite{ssd} & 1331M & 0.27800 & 0.2529 & 0.9245 \\
     & Ours & 1316M & \textbf{0.27873} & \textbf{0.2644} & \textbf{0.9437} \\ \cline{2-6} 
     & Vega\cite{ssd} & 745M & 0.27598 & 0.2412 & 0.7790 \\
     & Ours & 729M & \textbf{0.27758} & \textbf{0.2421} & \textbf{0.8210} \\ \hline
    \multirow{5}{*}{\begin{tabular}[c]{@{}c@{}}SDXL\\ Teacher: SDXL-Base-1.0\cite{sdxl_base}\end{tabular}} & Teacher & 2567M & 0.27213 & 0.2459 & 0.5187 \\ \cline{2-6} 
     & KOALA\cite{koala} & 1161M & \textbf{0.26963} & 0.2101 & 0.2604 \\
     & Ours & 1142M & \textbf{0.26968} & \textbf{0.2208} & \textbf{0.3281} \\ \cline{2-6} 
     & KOALA\cite{koala} & 782M & 0.26893 & 0.2045 & 0.2302 \\
     & Ours & 764M & \textbf{0.26913} & \textbf{0.2083} & \textbf{0.2603} \\ \hline
    \multirow{5}{*}{\begin{tabular}[c]{@{}c@{}}SDM-v1.5\\ Teacher: RealisticVision-v4.0\cite{rv40}\end{tabular}} & Teacher & 860M & 0.27803 & 0.2399 & 0.7422 \\ \cline{2-6} 
     & BK-SDM\cite{bk_sdm, small_sd} & 579M & 0.27450 & 0.1999 & 0.5619 \\
     & Ours & 578M & \textbf{0.27575} & \textbf{0.2136} & \textbf{0.5807} \\ \cline{2-6} 
     & BK-SDM\cite{bk_sdm, tiny_sd} & 323M & 0.27030 & 0.1780 & 0.3142 \\
     & Ours & 320M & \textbf{0.27078} & \textbf{0.1782} & \textbf{0.3201} \\ \hline
    \end{tabular}
}
\caption{Comparisons with the state-of-the-art methods.}
\label{tab_comp_sota}
\end{table*}

\section{Experiments}

\subsection{Implementation} \label{sec_implementation}
In this subsection, we mainly elaborate on the basic implementation settings. Further details are in the supplementary.

\textbf{Model selection.}
Experiments in Sec~\ref{sec_ablation_additivity}, Sec~\ref{sec_ablation_criteria} and Sec~\ref{sec_ablation_distill} are conducted on 512$\times$512 resolution for fast validation, using the ProtoVisionXL-v6.2.0\cite{protovision} for SDXL model since the officially released SDXL-Base-1.0\cite{sdxl_base} renders anomalous images around the 512$\times$512 resolution, and using the officially released 
stable-diffusion-v1-5\cite{sd_v1_5} for SDM-v1.5 model.
For experiments in Sec~\ref{sec_comp_sota}, we use the same teacher model of each comparative method.

\textbf{Dataset.}
For the calibration dataset used for pruning, we use a randomly sampled 1K subset of LAION-2B\cite{laion_2b}.
For the ablation studies in Sec~\ref{sec_ablation_criteria} and Sec~\ref{sec_ablation_distill}, we use a randomly sampled 0.34M subset of LAION-2B\cite{laion_2b}.
For the comparisons of the state-of-the-art methods in Sec.\ref{sec_comp_sota}, we use the same datasets of the comparative methods or turn to the datasets with lower quality and quantity if their reported datasets are hard to reproduce.
The selections of datasets are detailed in the supplementary.

\textbf{Evaluation metric.}
Although it has been a common practice to evaluate generative T2I models by FID\cite{fid} and CLIP score\cite{clipscore}, recent studies\cite{fid_is_bad_1, pick_a_pic, sdxl, hps, hps_v2} showed that these two metrics are poorly correlated with visual aesthetics and human preference.
Therefore, we adopted 3 advanced metrics to evaluate the model's comprehensive performance.
We use HPS v2\cite{hps_v2}, PickScore\cite{pick_a_pic}, and ImageReward\cite{image_reward} to evaluate the general visual quality of the generated images and the text-image consistency.
We calculate these 3 metrics on their own benchmark datasets, i.e., HPS v2 is calculated on its 3.2K benchmark dataset, ImageReward is calculated on a randomly sampled 3K subset of ImageRewardDB\cite{image_reward_db}, and PickScore is calculated on a randomly sampled 3K subset of Pick-a-Pic v1\cite{pick_a_pic_db}.

\subsection{Comparison with Other Methods} \label{sec_comp_sota}
Up to now, there are three handcrafted layer removal methods for SDMs.
KOALA\cite{koala}, SSD and Vega\cite{ssd} are on SDXL, and BK-SDM\cite{bk_sdm} can be applied on SDM-v1 or SDM-v2.
We compare our LAPTOP-Diff with these three methods on the corresponding models.
We use the SDXL-Base-1.0\cite{sdxl_base} for the comparison with KOALA 1.16B and 782M, the RealisticVision-v4.0\cite{rv40} for the comparison with the more advanced BK-SDM\cite{bk_sdm} implementation by Segmind, i.e., 579M small-sd\cite{small_sd} and 323M tiny-sd\cite{tiny_sd}.
For the comparison with 1.3B SSD\cite{ssd} and 745M Vega\cite{ssd}, we only use one of the three teacher models they used in the multi-teacher distilling strategy, i.e., the ZavychromaXL-v1.0\cite{zavy}.
For all the evaluations of the comparative methods, we used their released model weights.

Tab.\ref{tab_comp_sota} showed the comparison results with the state-of-the-art compressed SDM models.
Our proposed LAPTOP-Diff achieved the most advanced performance.
The visual comparisons with SSD\cite{ssd} and Vega\cite{ssd} are shown in Fig.\ref{fig_visual_comp}, and more visual comparisons with other methods are shown in the supplementary.
We can observe that our method achieved better visual results on different prompts compared with other methods.
Notably, it is also observed that our around 50\% compressed SDXL model can achieve almost the same visual quality of the original model.

\subsection{Validation of Additivity Property} \label{sec_ablation_additivity}
\begin{figure}[t]
	\centering
	\subfloat[Criterion: Output Loss. On SDXL.]{\includegraphics[width=0.4\columnwidth]{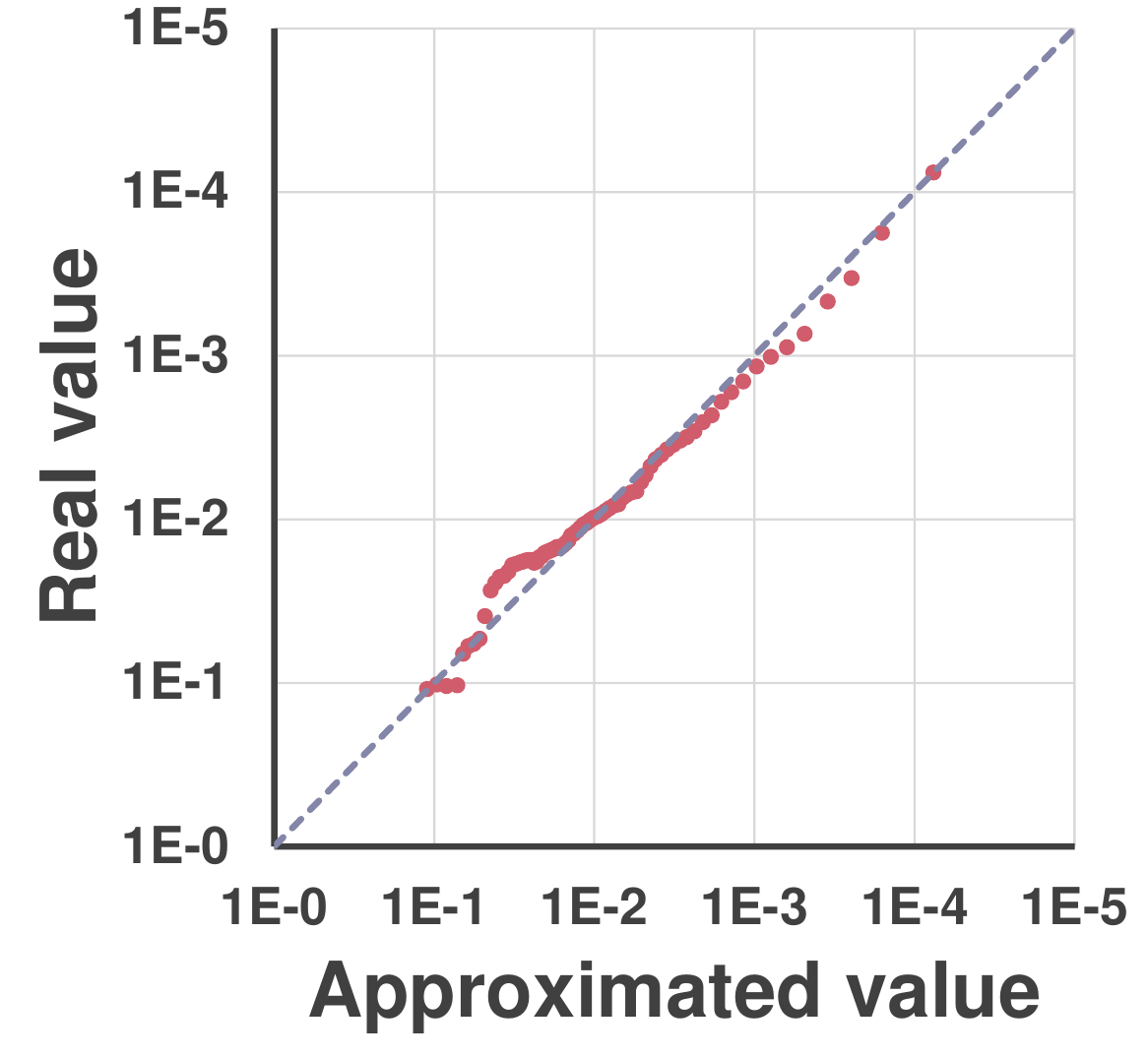}}\hspace{25pt}
	\subfloat[Criterion: Output Loss. On SDM-v1.5.]{\includegraphics[width=0.4\columnwidth]{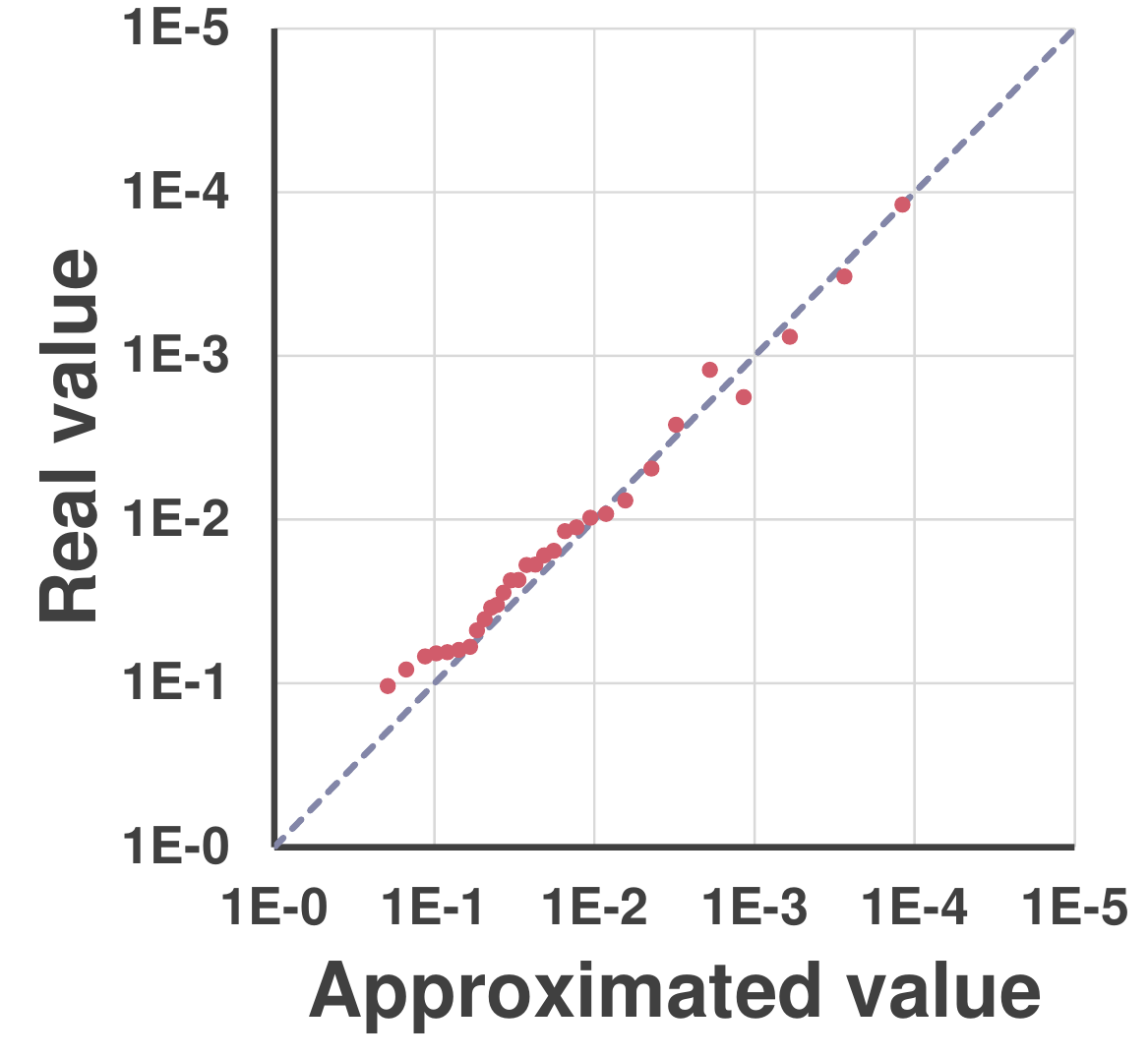}}\\
	\subfloat[Criterion: $\Delta$Task Loss. On SDXL.]{\includegraphics[width=0.4\columnwidth]{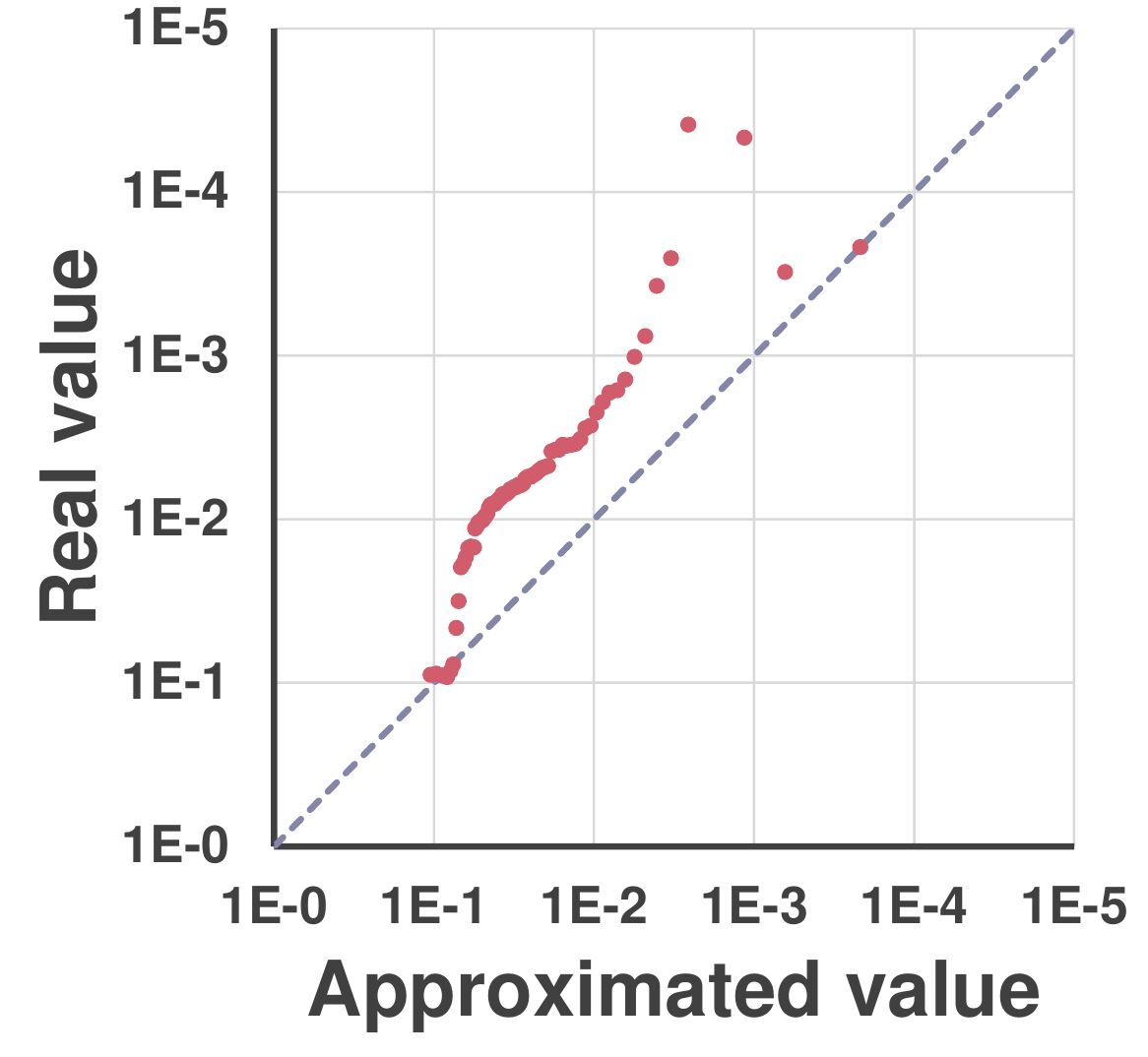}}\hspace{25pt}
	\subfloat[Criterion: $\Delta$Task Loss. On SDM-v1.5.]{\includegraphics[width=0.4\columnwidth]{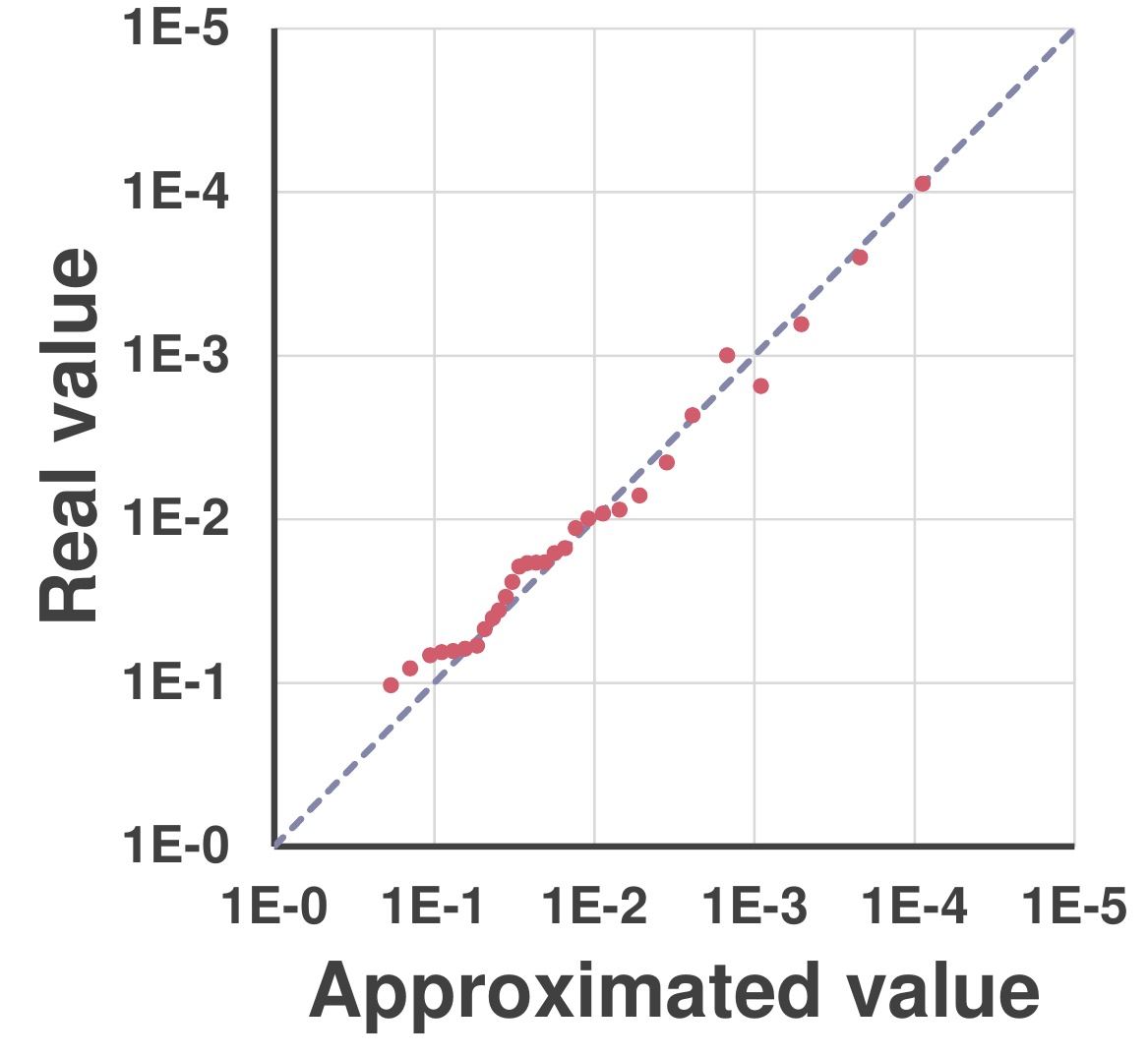}}\\
	\subfloat[Criterion: $\Delta$CLIP score. On SDXL.]{\includegraphics[width=0.4\columnwidth]{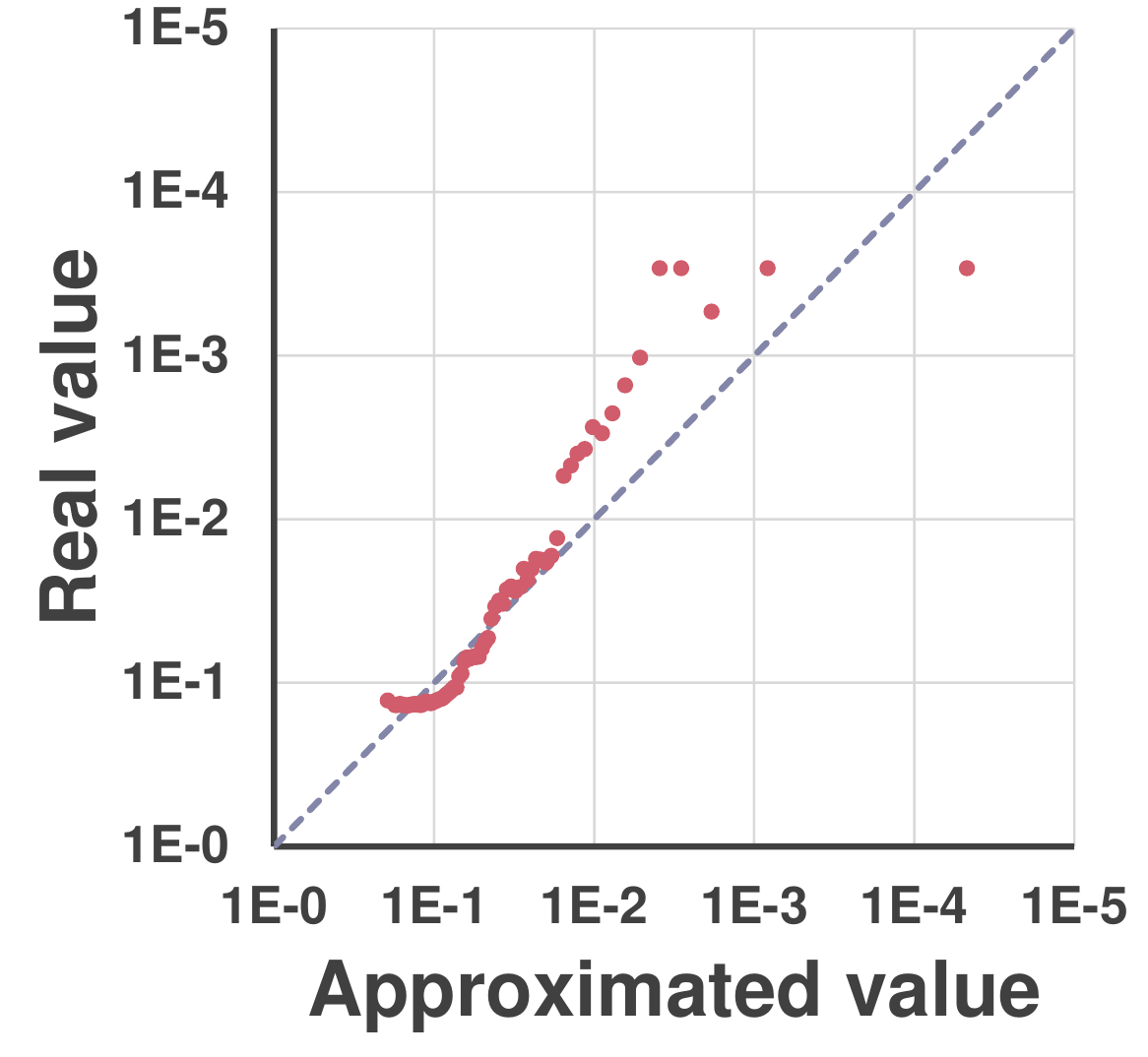}}\hspace{25pt}
	\subfloat[Criterion: $\Delta$CLIP score. On SDM-v1.5.]{\includegraphics[width=0.4\columnwidth]{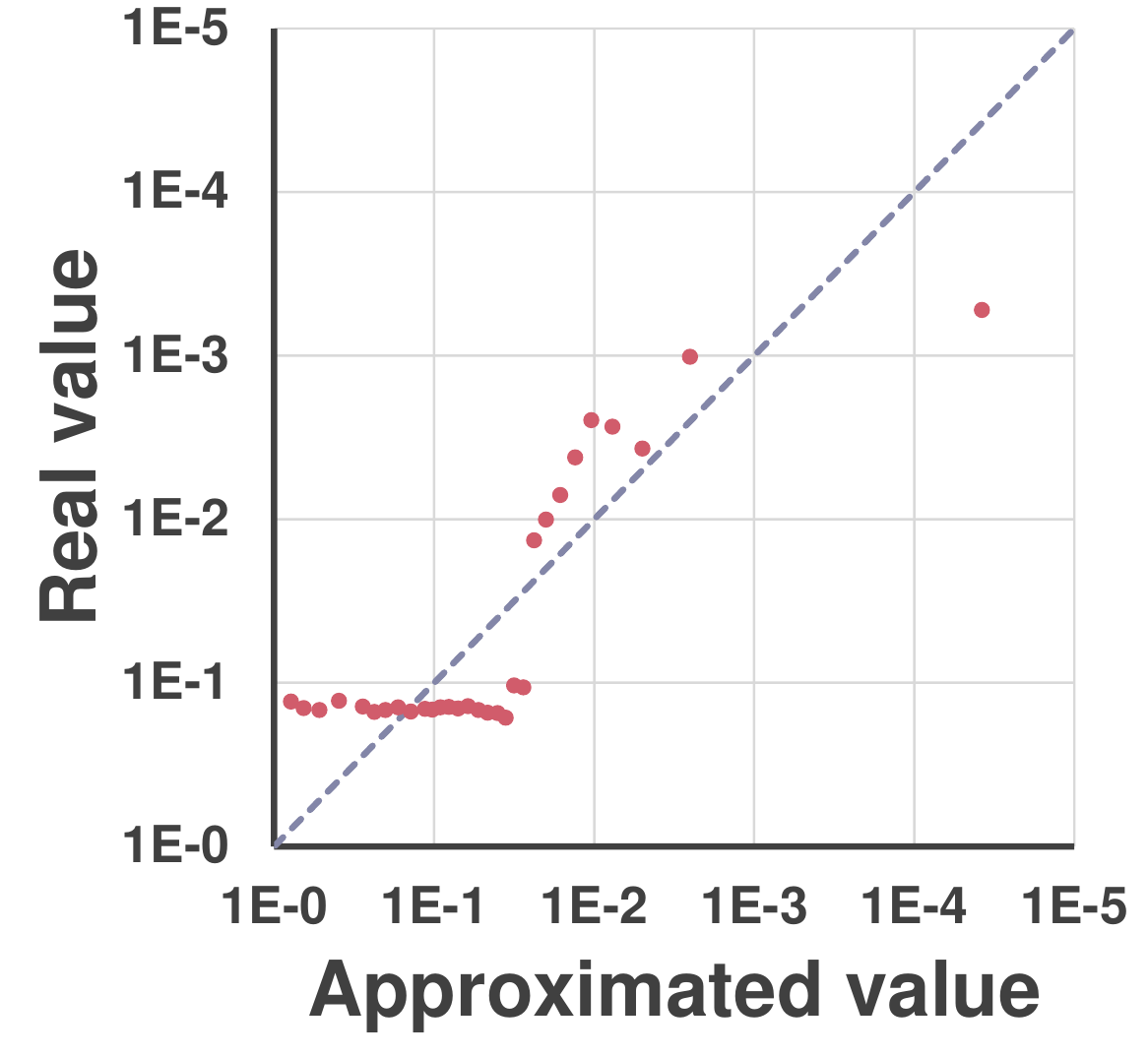}}
	\caption{Validations of the additivity properties of different pruning criteria on SDXL and SDM-v1.5. Each point represents an observed value pair of the approximated criterion and the real criterion.}
\label{fig_validate_additivity}
\end{figure}

Since the final surrogate objective (\ref{objective_3}) is the approximation of an upper bound of the original objective (\ref{objective_1}), it is necessary to examine the precision of the approximation.
We simulate the values of the approximated criterion (\ref{objective_3}), and the real criterion (\ref{objective_1}), by choosing different $m$ and $R^m$, covering the pruning ratio from $0\%$ to $90\%$.
We also conduct the same experiments on pruning criteria $\Delta$task loss and $\Delta$CLIP score.
As shown in Fig.\ref{fig_validate_additivity}, our output loss criterion significantly satisfies the additivity property across different models even under an extreme pruning ratio of $90\%$, where almost all observation points are located near 
the identity line. Therefore, our final surrogate objective (\ref{objective_3}) is an excellent approximation of the original objective (\ref{objective_1}). On the other hand, other criteria fail to consistently satisfy the additivity property.

While our output loss criterion holds the strongest additivity property across different models, the $\Delta$CLIP score holds the second strongest additivity on SDXL, and the $\Delta$task loss holds the second strongest additivity on SDM-v1.5.

\subsection{Ablation of Pruning Criteria} \label{sec_ablation_criteria}
Experiments in Sec~\ref{sec_ablation_additivity} showed that our output loss criterion significantly and consistently satisfies the additivity property while others fail.
Furthermore, we evaluated the pruning performance using different pruning criteria including the pruning criteria we constructed and two baseline pruning criteria, i.e., the magnitude metric and the first-order Taylor-expansion of the loss function.
We first prune the SDXL and SDM-v1.5 using different layer pruning criteria and then retrain the pruned network by vanilla distillation for the same training iterations.
As shown in Tab.\ref{tab_ablation_criteria}, the output loss criterion attains the highest pruning performance among different pruning criteria on different models.

In addition to different pruning criteria, we also compared the layer pruning methods with handcrafted layer removal approaches.
For the experiments on SDXL and SDM-v1.5, we use the same handcrafted layer removal settings of SSD\cite{ssd} and BK-SDM\cite{bk_sdm} respectively.
Results in Tab.\ref{tab_ablation_criteria} showed that almost all the layer pruning approaches generally surpassed the handcrafted approaches except for the magnitude-based layer pruning is beaten by handcrafted approach SSD\cite{ssd} on SDXL.

It is worth notice that, among the three pruning criteria we constructed, i.e., output loss, $\Delta$task loss, and $\Delta$CLIP score, while the output loss achieves the best performance across different models, the $\Delta$CLIP score achieves the second best performance on SDXL, and the $\Delta$task loss achieves the second best performance on SDM-v1.5, aligned with their ranks on additivity in Sec.\ref{sec_ablation_additivity}.

\begin{table}[t]
\caption{Comparisons of different pruning criteria. In addition, comparisons with handcrafted layer removal approaches are included.}
\scalebox{0.8}{
    \begin{tabular}{ccccc}
    \hline
    \multicolumn{2}{c}{Method} & HPS v2$\uparrow$ & PickScore$\uparrow$ & ImageReward$\uparrow$ \\ \hline
    \multirow{7}{*}{\begin{tabular}[c]{@{}c@{}}SDXL\\ 50\% pruned\end{tabular}} & Teacher & 0.28085 & 0.2762 & 1.1769 \\ \cline{2-5} 
     & SSD\cite{ssd} & 0.27753 & 0.2445 & 0.8665 \\
     & Magnitude & 0.26888 & 0.1980 & 0.4755 \\
     & Taylor & \textbf{0.27925} & 0.2519 & 0.9290 \\
     & $\Delta$Task Loss & 0.27750 & 0.2493 & 0.9398 \\
     & $\Delta$CLIP Score & 0.27888 & 0.2537 & 0.9607 \\
     & Output Loss(Ours) & \textbf{0.27928} & \textbf{0.2559} & \textbf{0.9844} \\ \hline
    \multirow{7}{*}{\begin{tabular}[c]{@{}c@{}}SDM-v1.5\\ 33\% pruned\end{tabular}} & Teacher & 0.27023 & 0.1965 & 0.1442 \\ \cline{2-5} 
     & BK-SDM\cite{bk_sdm} & 0.25988 & 0.1148 & -0.5698 \\
     & Magnitude & 0.26160 & 0.1464 & -0.3112 \\
     & Taylor & 0.26318 & 0.1481 & -0.3056 \\
     & $\Delta$Task Loss & 0.26390 & 0.1367 & -0.3247 \\
     & $\Delta$CLIP Score & 0.26215 & 0.1351 & -0.3715 \\
     & Output Loss(Ours) & \textbf{0.26625} & \textbf{0.1517} & \textbf{-0.2189} \\ \hline
    \end{tabular}
}
\label{tab_ablation_criteria}
\end{table}

\begin{table}
\caption{Comparisons between vanilla distillation and our normalized distillation.}
\scalebox{0.8}{
    \begin{tabular}{ccccc}
    \hline
    \multicolumn{2}{c}{Method} & HPS v2$\uparrow$ & PickScore$\uparrow$ & ImageReward$\uparrow$ \\ \hline
    \multirow{3}{*}{\begin{tabular}[c]{@{}c@{}}SDXL\\ 50\% pruned\end{tabular}} & Teacher & 0.28085 & 0.2762 & 1.1769 \\ \cline{2-5} 
     & Vanilla KD & 0.27933 & 0.2562 & 0.9852 \\
     & Normed KD(Ours) & \textbf{0.27963} & \textbf{0.2564} & \textbf{1.0164} \\ \hline
    \multirow{3}{*}{\begin{tabular}[c]{@{}c@{}}SDM-v1.5\\ 33\% pruned\end{tabular}} & Teacher & 0.27023 & 0.1965 & 0.1442 \\ \cline{2-5} 
     & Vanilla KD & 0.26780 & 0.1596 & -0.0573 \\
     & Normed KD(Ours) & \textbf{0.26810} & \textbf{0.1642} & \textbf{-0.0325} \\ \hline
    \end{tabular}
}
\label{tab_ablation_distill}
\end{table}

\subsection{Ablation of Knowledge Distillation} \label{sec_ablation_distill}
We validate the effectiveness of our normalized feature distillation through ablation studies.
We prune the SDXL and SDM-v1.5 by our one-shot layer pruning method and retrain the pruned network using vanilla distillation or our normalized distillation for the same training iterations.
As shown in Tab.\ref{tab_ablation_distill}, our method achieved great performance improvements on both SDXL and SDM-v1.5.

\subsection{Pruning Analysis} \label{sec_pruning_analysis}
We analyzed the pruned architectures of SDXL's U-Net at a pruning ratio of 50\% and SDM-v1.5's U-Net at a pruning ratio of 33\% as examples.
As Tab.\ref{tab_pruned_arch} shows, for both SDXL and SDM-v1.5, many layers in and near the Mid stage are considered less important and thus were pruned.
This observation is consistent with the previous handcrafted layer removal approaches\cite{bk_sdm, ssd, koala}.
However, undiscovered by the previous handcrafted layer removal approaches, we observed that more layers are pruned at the Dn stages than the Up stages.
For SDXL, we observed that 18 of the 30 layers were pruned in the Dn stages, but only 10 of the 45 were pruned in the Up stages.
The same phenomenon is observed on SDM-v1.5 that 4 of the 14 layers were pruned in the Dn stages, while only 1 of the 21 were pruned in the Up stages.
This observation is consistent with other studies\cite{faster_diffusion, deep_cache} that the Dn stage, i.e., the encoder of U-Net, is less important than the other parts of the U-Net.

\begin{table}
\caption{Pruned U-Net architectures of SDXL and SDM-v1.5, using SDXL-Base-1.0\cite{sdxl_base} for SDXL at a total pruning ratio of 50\% and stable-diffusion-v1-5\cite{sd_v1_5} for SDM-v1.5 at a total pruning ratio of 33\% as examples.}
\scalebox{0.8}{
    \begin{threeparttable}
    \begin{tabular}{ccccccc}
    \hline
    Model & Stage & $\#R_{ori}$ & $\#R_{prn}$ & $\#T_{ori}$ & $\#T_{prn}$ & P-Ratio \\ \hline
    \multirow{9}{*}{\begin{tabular}[c]{@{}c@{}}SDXL\\ 50\% pruned\end{tabular}} & Dn 1 & 2 & 0 & 0 & 0 & 0\% \\
     & Dn 2 & 2 & 0 & 4 & 0 & 0\% \\
     & Dn 3 & 2 & 0 & 20 & 18 & 83\% \\
     & Mid & 2 & 0 & 10 & 8 & 67\% \\
     & Up 1 & 3 & 0 & 30 & 10 & 29\% \\
     & Up 2 & 3 & 0 & 6 & 0 & 0\% \\
     & Up 3 & 3 & 0 & 0 & 0 & 0\% \\ \cline{2-7} 
     & Dn Total & 6 & 0 & 24 & 18 & 76\% \\
     & Up Total & 9 & 0 & 36 & 10 & 26\% \\ \hline
    \multirow{11}{*}{\begin{tabular}[c]{@{}c@{}}SDM-v1.5\\ 33\% pruned\end{tabular}} & Dn 1 & 2 & 0 & 2 & 0 & 0\% \\
     & Dn 2 & 2 & 0 & 2 & 0 & 0\% \\
     & Dn 3 & 2 & 0 & 2 & 2 & 50\% \\
     & Dn 4 & 2 & 2 & 0 & 0 & 100\% \\
     & Mid & 2 & 2 & 1 & 1 & 100\% \\
     & Up 1 & 3 & 0 & 0 & 0 & 0\% \\
     & Up 2 & 3 & 0 & 3 & 1 & 13\% \\
     & Up 3 & 3 & 0 & 3 & 0 & 0\% \\
     & Up 4 & 3 & 0 & 3 & 0 & 0\% \\ \cline{2-7} 
     & Dn Total & 8 & 2 & 6 & 2 & 53\% \\
     & Up Total & 12 & 0 & 9 & 1 & 7\% \\ \hline
    \end{tabular}
    \begin{tablenotes}
    \item $\#R_{ori}$: number of residual layers before pruning; $\#R_{prn}$: number of pruned residual layers; $\#T_{ori}$: number of transformer layers before pruning; $\#T_{prn}$: number of pruned transformer layers; P-Ratio: pruning ratio in a stage.
    \end{tablenotes}
    \end{threeparttable}
}
\label{tab_pruned_arch}
\end{table}

\section{Conclusion}
In this work, we proposed the layer pruning and normalized distillation for compressing diffusion models(LAPTOP-Diff).
We introduced the layer pruning approach to achieve automation, scalability, and better performance, and proposed an effective one-shot pruning criterion, i.e., the output loss criterion, whose effectiveness is ensured by its good additivity property.
We further alleviated the imbalance issue of the previous distillation-based retraining using our proposed normalized feature distillation.
Using the proposed LAPTOP-Diff, we compressed the SDMs for the most advanced performance.


\bibliographystyle{ACM-Reference-Format}

\clearpage
\appendix

\section{Solving the Variant of 0-1 Knapsack Problem} \label{supplementaryknapsack_problem}
The optimization problem (\ref{objective_3}) in the main paper is a variant of 0-1 Knapsack problem.
The $\mathbb{E}||\epsilon(l_i)-\epsilon_{ori}||_2^2$, $params(l_i)$ and $P$ correspond to the value of the item, weight of the item and the knapsack capacity respectively.
The objective is to find a subset of items that has a minimal total value while the total weight is no less than $P$.
Similar to the original 0-1 Knapsack problem, this variant problem can also be solved by the dynamic programming algorithm or can be approximately solved by the greedy algorithm.
We give the pseudocodes of the greedy algorithm and the dynamic programming algorithm as Algorithm.\ref{algo_knapsack_greedy} and Algorithm.\ref{algo_knapsack_dp} respectively.

\begin{algorithm}
\caption{Solution to the variant of 0-1 Knapsack problem using greedy algorithm}
\label{algo_knapsack_greedy}
\begin{algorithmic}[1]
\renewcommand{\algorithmicrequire}{\textbf{Input:}}
\renewcommand{\algorithmicensure}{\textbf{Output:}}
\Require{$values$ array, $weights$ array, the $knapsack\_capacity$ $P$ }
\Ensure{An index set $S$ of items that has a minimal total value while the total weight is no less than $P$}
\State $n$ = $length(values)$
\State $S = \emptyset$
\State $k = 0$
\State $p = 0$ \Comment{current cumulative weight of the items in the bag}
\While{$p<P$}
    \State $i =$ index of the item that has the $k$-th smallest $value$
    \State $p = p + weights[i]$
    \State $S = S \cup \{i\}$ \Comment{the $i$-th item is in the bag}
    \State $k = k + 1$
\EndWhile \\
\Return $S$
\end{algorithmic}
\end{algorithm}

\begin{algorithm}
\caption{Solution to the variant of 0-1 Knapsack problem using dynamic programming algorithm}
\label{algo_knapsack_dp}
\begin{algorithmic}[1]
\renewcommand{\algorithmicrequire}{\textbf{Input:}}
\renewcommand{\algorithmicensure}{\textbf{Output:}}
\Require{$values$ array, $weights$ array, the $knapsack\_capacity$ $P$ }
\Ensure{An index set $S$ of items that has a minimal total value while the total weight is no less than $P$}

\Statex \textbf{Search for a minimal objective value:}
\State $n = length(values)$
\State $dp =$ zero array of shape $(n, P+1)$
\For{$j =$ $0$ to $P$}
    \State \textbf{if} $weight[0] < j$ \textbf{then} $dp[0, j] = infinite$
    \State \textbf{else} $dp[0, j] = value[0]$
\EndFor

\For{$i =$ $0$ to $n-1$}
    \For{$j =$ $P$ to $0$}
        \State \textbf{if} $weight[i] < j$ \textbf{then} $dp[i, j] = min(dp[i - 1, j], dp[i - 1, j - weight[i]] + value[i])$
        \State \textbf{else} $dp[i, j] = min(dp[i - 1, j], value[i])$
        \State \textbf{if} $sum(weight[:i + 1]) < j$ \textbf{then} $dp[i, j] = infinity$
    \EndFor
\EndFor

\Statex
\Statex \textbf{Get the solution to the minimal objective:}
\State $S = \emptyset$
\def\Tab{\hspace{\algorithmicindent}}
\State \textbf{def} \textbf{DFS}(i, j):
    \State\Tab \textbf{if} $j = 0$ \textbf{then} \textbf{Return}
    \State\Tab \textbf{if} $i = 0$ \textbf{and} $weight[0] >= j$ \textbf{then}
        \State\Tab\Tab $S = S \cup \{i\}$ \Comment{the $i$-th item is in the bag}
        \State\Tab\Tab \textbf{Return}
    \State\Tab \textbf{if} $weight[i] < j$ \textbf{then}
        \State\Tab\Tab \textbf{if} $dp[i, j] = dp[i - 1, j - weight[i]] + value[i]$ \textbf{then}
            \State\Tab\Tab\Tab $S = S \cup \{i\}$ \Comment{the $i$-th item is in the bag}
            \State\Tab\Tab\Tab \textbf{DFS}($i - 1, j - weight[i]$) \Comment{depth-first search}
        \State\Tab\Tab \textbf{else} \textbf{DFS}($i - 1, j$) \Comment{depth-first search}
    \State\Tab \textbf{else}
        \State\Tab\Tab \textbf{if} $dp[i, j] = value[i]$ \textbf{then}
            \State\Tab\Tab\Tab $S = S \cup \{i\}$ \Comment{the $i$-th item is in the bag}
            \State\Tab\Tab\Tab \textbf{Return}
        \State\Tab\Tab \textbf{else} \textbf{DFS}($i - 1, j$) \Comment{depth-first search}
\State \textbf{DFS}($n-1, P$) \Comment{start the recursion} \\
\Return $S$
\end{algorithmic}
\end{algorithm}

\begin{table*}[!t]
\scalebox{0.74}{
    \begin{tabular}{ccccccc}
    \hline
    Section & Model & Dataset & Resolution & Total Batch Size & Iterations & Learning Rate \\ \hline
    \multirow{3}{*}{Sec~\ref{sec_comp_sota}} & ZavychromaXL v1.0\cite{zavy} for the comparison with   SSD\cite{ssd} and Vega\cite{ssd} & 8M LAION Aesthetics V2 6+\cite{laion_ae} & 1024$\times$1024 & \multirow{3}{*}{128} & 100K and 240K & 1E-5 \\
     & SDXL-Base-1.0\cite{sdxl_base}   for the comparison with KOALA\cite{koala} & 8M LAION Aesthetics V2 6+\cite{laion_ae} & 1024$\times$1024 &  & 125K & 1E-5 \\
     & RealisticVision-v4.0\cite{rv40}   for the comparison with BK-SDM\cite{bk_sdm} & 0.34M subset of LAION-2B\cite{laion_2b} & 512$\times$512 &  & 120K & 1E-4 \\ \hline
    \multirow{2}{*}{Sec~\ref{sec_ablation_additivity}} & ProtoVisionXL-v6.2.0\cite{protovision} for SDXL & \multirow{2}{*}{1K subset of LAION-2B\cite{laion_2b}} & \multirow{2}{*}{512$\times$512} & \multirow{2}{*}{-} & \multirow{2}{*}{-} & \multirow{2}{*}{-} \\
     & stable-diffusion-v1-5\cite{sd_v1_5}   for SDM-v1.5 &  &  &  &  &  \\ \hline
    \multirow{2}{*}{Sec~\ref{sec_ablation_criteria}} & ProtoVisionXL-v6.2.0\cite{protovision} for SDXL & \multirow{2}{*}{0.34M subset of LAION-2B\cite{laion_2b}} & \multirow{2}{*}{512$\times$512} & 512 & 15K & \multirow{2}{*}{1E-5} \\
     & stable-diffusion-v1-5\cite{sd_v1_5}   for SDM-v1.5 &  &  & 128 & 12K &  \\ \hline
    \multirow{2}{*}{Sec~\ref{sec_ablation_distill}} & ProtoVisionXL-v6.2.0\cite{protovision} for SDXL & \multirow{2}{*}{0.34M subset of LAION-2B\cite{laion_2b}} & \multirow{2}{*}{512$\times$512} & 512 & 30K & \multirow{2}{*}{1E-5} \\
     & stable-diffusion-v1-5\cite{sd_v1_5}   for SDM-v1.5 &  &  & 128 & 60K &  \\ \hline
    \end{tabular}
}
\caption{Implementation details of the experiments in each section of the main paper.}
\label{tab_implementation}
\end{table*}

\section{Derivations of Other Pruning Criteria} \label{supplementaryderivation}

\subsection{Derivation of $\Delta$task loss} \label{t_derivation}
Let $\mathcal{L}_{ori}$ denote the task loss of the original network, $\mathcal{L}(r_1, r_2, \dots, r_m)$ denote the task loss of the network where layers $r_1, r_2, \dots, r_m$ are removed.
We have the combinatorial optimization objective:
\begin{equation}
\begin{aligned}
\mathop{\min}_{r_1, \ldots, r_m}
\mathbb{E}|\mathcal{L}&(r_1, \ldots, r_m)-\mathcal{L}_{ori}| \\
s.t.\ \{r_1, \ldots, r_m\} &\subset L^n,\ \sum_{i=1}^{m}params(r_i) \geq P.
\end{aligned}
\label{t_objective_1}
\end{equation}
According to the triangle inequality, we have an upper bound of the objective (\ref{t_objective_1}):
\begin{equation}
\begin{aligned}
\mathbb{E}&|\mathcal{L}(r_1, \ldots, r_m)-\mathcal{L}_{ori}| \\
\leq&\mathbb{E}|\mathcal{L}(r_1)-\mathcal{L}_{ori}| \\
&+\mathbb{E}|\mathcal{L}(r_1, r_2)-\mathcal{L}(r_1)| \\
&+\ldots \\
&+\mathbb{E}|\mathcal{L}(r_1, \ldots, r_m)-\mathcal{L}(r_1, \ldots, r_{m-1})|.
\end{aligned}
\label{t_inequality_1}
\end{equation}
Similar to the assumption (\ref{approximation_1}) in the main paper, we have the assumption:
\begin{equation}
\begin{aligned}
&\mathbb{E}|\mathcal{L}(r_1, \ldots, r_{i-1}, r_i)-\mathcal{L}(r_1, \ldots, r_{i-1})| \\
&\approx\mathbb{E}|\mathcal{L}(r_1, \ldots, r_{i-2}, r_i)-\mathcal{L}(r_1, \ldots, r_{i-2})| \\
&\approx\dots \\
&\approx\mathbb{E}|\mathcal{L}(r_1, r_i)-\epsilon(r_1)| \\
&\approx\mathbb{E}|\mathcal{L}(r_i)-\mathcal{L}_{ori}|.
\end{aligned}
\label{t_approximation_1}
\end{equation}
However, different from assumption (\ref{approximation_1}), the assumption (\ref{t_approximation_1}) may not be satisfied well across different models according to the Validation of Additivity Property section in the main paper.

With the approximation (\ref{t_approximation_1}) and the upper bound (\ref{t_inequality_1}), we have the surrogate objective:
\begin{equation}
\begin{aligned}
\mathop{\min}_{r_1, \ldots, r_m}
\sum_{i=1}^{m}&\mathbb{E}|\mathcal{L}(r_i)-\mathcal{L}_{ori}| \\
s.t.\ \{r_1, \ldots, r_m\} &\subset L^n,\ \sum_{i=1}^{m}params(r_i) \geq P.
\end{aligned}
\label{t_objective_2}
\end{equation}
Here, $\mathbb{E}|\mathcal{L}(r_i)-\mathcal{L}_{ori}|$ is the $\Delta$task loss between the original network and the network where only layer $r_i$ is removed.

\subsection{Derivation of $\Delta$CLIP score} \label{c_derivation}
Let $\mathcal{C}_{ori}$ denote the CLIP score of the original network, $\mathcal{C}(r_1, r_2, \dots, r_m)$ denote the CLIP score of the network where layers $r_1, r_2, \dots, r_m$ are removed.
The form of the derivation of $\Delta$CLIP score is the same as the derivation of $\Delta$task loss. Simply replace the $\mathcal{L}$ and $\mathcal{L}_{ori}$ appear in the Supplementary.\ref{t_derivation} with $\mathcal{C}$ and $\mathcal{C}_{ori}$ respectively, and we can get the final objective:
\begin{equation}
\begin{aligned}
\mathop{\min}_{r_1, \ldots, r_m}
\sum_{i=1}^{m}&\mathbb{E}|\mathcal{C}(r_i)-\mathcal{C}_{ori}| \\
s.t.\ \{r_1, \ldots, r_m\} &\subset L^n,\ \sum_{i=1}^{m}params(r_i) \geq P.
\end{aligned}
\label{c_objective}
\end{equation}
Here, $\mathbb{E}|\mathcal{C}(r_i)-\mathcal{C}_{ori}|$ is the $\Delta$CLIP score between the original network and the network where only layer $r_i$ is removed.

\section{Further Implementation Details} \label{supplementaryimplementation}
In addition to the basic implementation information in the main paper, further details are summarised in Tab.\ref{tab_implementation}.

All the experiments are conducted with the Diffusers library using FP16 precision. We use the same VAEs and text encoders of the original SDXL and SDM-v1.5, except for the sdxl-vae-fp16-fix\cite{sdxl_vae_fp16} VAE checkpoint for SDXL model since the SDXL's original VAE is incompatible with FP16 precision computation.

For training, we use the AdamW optimizer with constant learning rates shown in Tab.\ref{tab_implementation} while other hyperparameters of the optimizer maintain default.

For inference, we use the DDIM scheduler\cite{ddim} for SDXL, and DPM-Solver++\cite{dpm, dpm++} for SDM-v1.5, all with 25 denoising steps and the default 7.5 classifier-free guidance scale\cite{cfg}.

\section{Latency Analysis}
Tab.\ref{tab_latency} shows the latencies of the original models and our compressed models.
Our 50\% and 72\% compressed SDXL U-Nets are 33\% and 48\% faster than the original U-Net, and our 33\% and 63\% compressed SDM-v1.5 U-Nets are 17\% and 31\% faster than the original U-Net.
\begin{table}[!t]
\scalebox{0.98}{
    \begin{tabular}{ccccc}
    \hline
    Resolution & Model & U-Net(1) & U-Net(25) & Whole \\ \hline
    \multirow{3}{*}{$1024\times1024$} & SDXL 2567M & 0.113s & 2.83s & 3.25s \\
     & Ours 1316M & 0.076s & 1.90s & 2.33s \\
     & Ours 729M & 0.059s & 1.48s & 1.89s \\ \hline
    \multirow{3}{*}{$512\times512$} & SDM-v1.5 860M & 0.028s & 0.70s & 0.77s \\
     & Ours 578M & 0.023s & 0.58s & 0.68s \\
     & Ours 320M & 0.019s & 0.48s & 0.56s \\ \hline
    \end{tabular}
}
\caption{Latency analysis of the original models and our compressed models. U-Net(1) for the latency of a single denoising step, U-Net(25) for 25 denoising steps, and Whole for the whole generating process of an image in 25 denoising steps, including text encoder, U-Net, and VAE decoder computations. Latencies are measured on NVIDIA A100 GPU.}
\label{tab_latency}
\end{table}

\section{Visual Comparisons with Other Methods} \label{supplementaryvisual_comp}
\begin{figure*}[t]
	\centering
    \begin{minipage}{0.1\textwidth}
    \quad
    \end{minipage}\hspace{10pt}
    \begin{minipage}{0.64\textwidth}
	\ \ SDXL-Base-1.0 \quad KOALA 1161M \quad\ \ Ours 1142M \quad\ \ KOALA 782M \qquad Ours 764M
    \end{minipage}\\
    
    \begin{minipage}{0.1\textwidth}
    a cute Shiba Inu head in a cabbage
    \end{minipage}\hspace{10pt}
    \begin{minipage}{0.64\textwidth}
	\subfloat{\includegraphics[width=\textwidth]{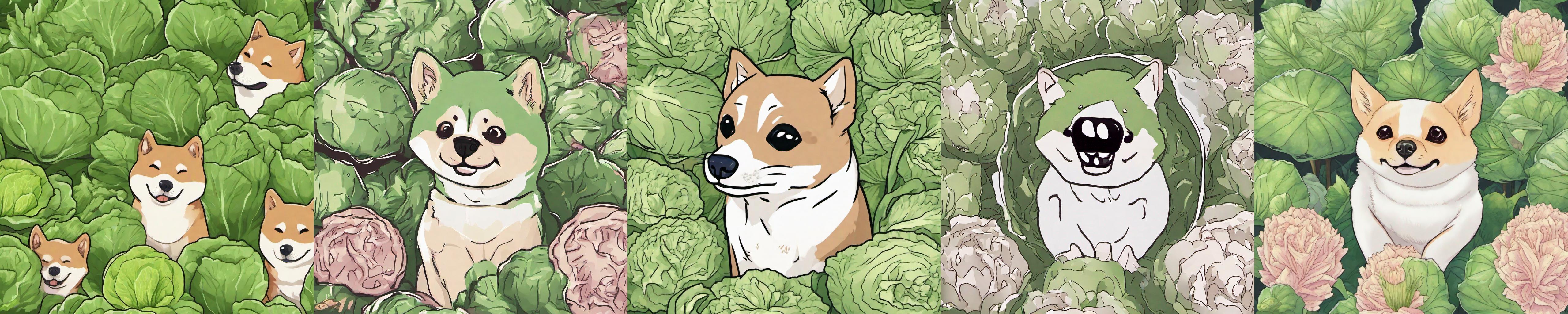}}
    \end{minipage}\\
    
    \begin{minipage}{0.1\textwidth}
    a giant cat carrying a small castle on its back, painting
    \end{minipage}\hspace{10pt}
    \begin{minipage}{0.64\textwidth}
	\subfloat{\includegraphics[width=\textwidth]{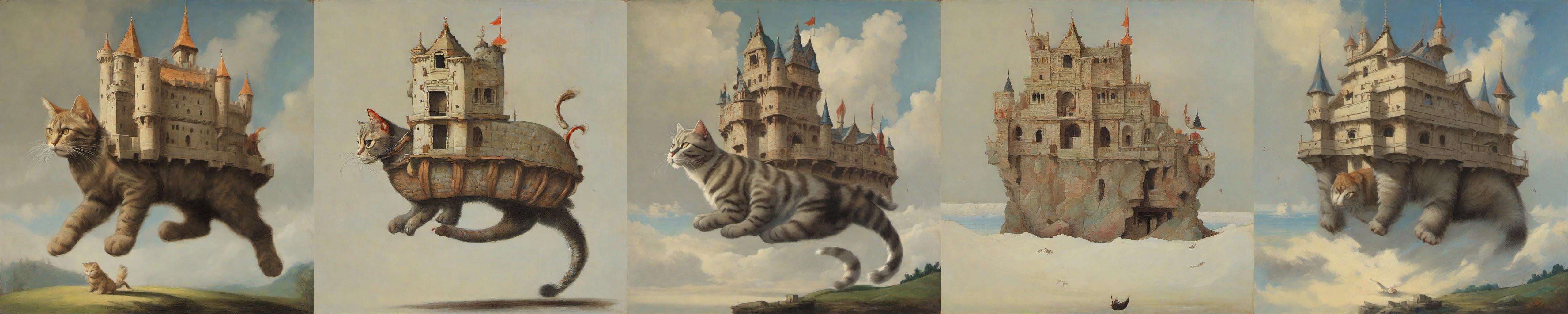}}
    \end{minipage}\\
    
    \begin{minipage}{0.1\textwidth}
    a classical guitar player at the concert
    \end{minipage}\hspace{10pt}
    \begin{minipage}{0.64\textwidth}
	\subfloat{\includegraphics[width=\textwidth]{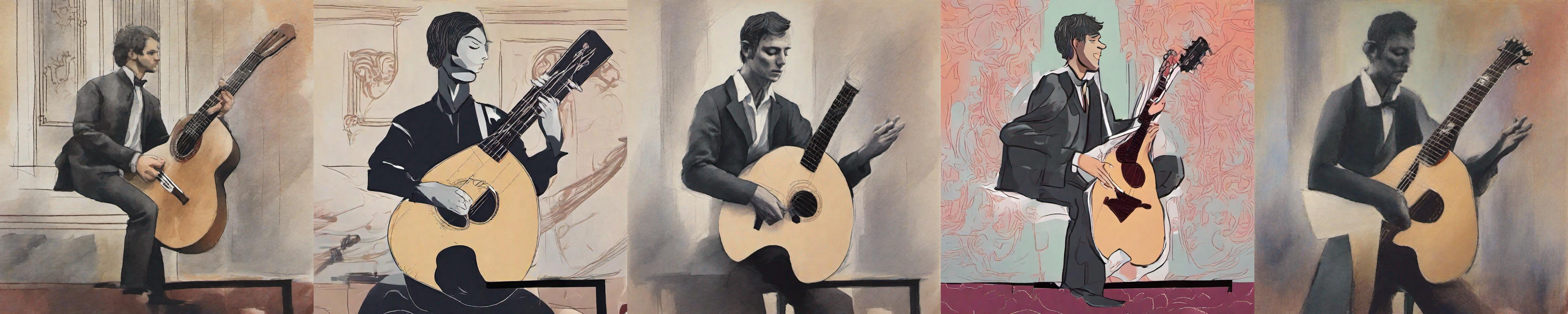}}
    \end{minipage}\\
    
    \begin{minipage}{0.1\textwidth}
    two tango dancers are dancing on the dim stage
    \end{minipage}\hspace{10pt}
    \begin{minipage}{0.64\textwidth}
	\subfloat{\includegraphics[width=\textwidth]{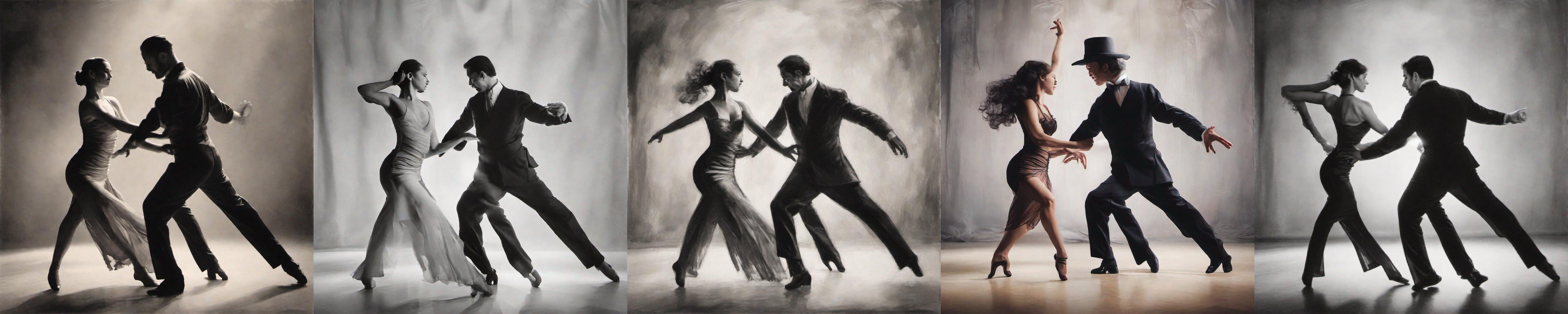}}
    \end{minipage}\\
\caption{Visual comparison with KOALA in DDIM 25 steps.}
\label{fig_visual_comp_koala}
\end{figure*}

\begin{figure*}[t]
	\centering
    \begin{minipage}{0.1\textwidth}
    \quad
    \end{minipage}\hspace{10pt}
    \begin{minipage}{0.64\textwidth}
	\qquad\ \ RV40 \qquad\quad BK-SDM 579M \quad\ \ Ours 578M \quad\ \ BK-SDM 323M \qquad Ours 320M
    \end{minipage}\\
    
    \begin{minipage}{0.1\textwidth}
    a cute Shiba Inu head in a cabbage
    \end{minipage}\hspace{10pt}
    \begin{minipage}{0.64\textwidth}
	\subfloat{\includegraphics[width=\textwidth]{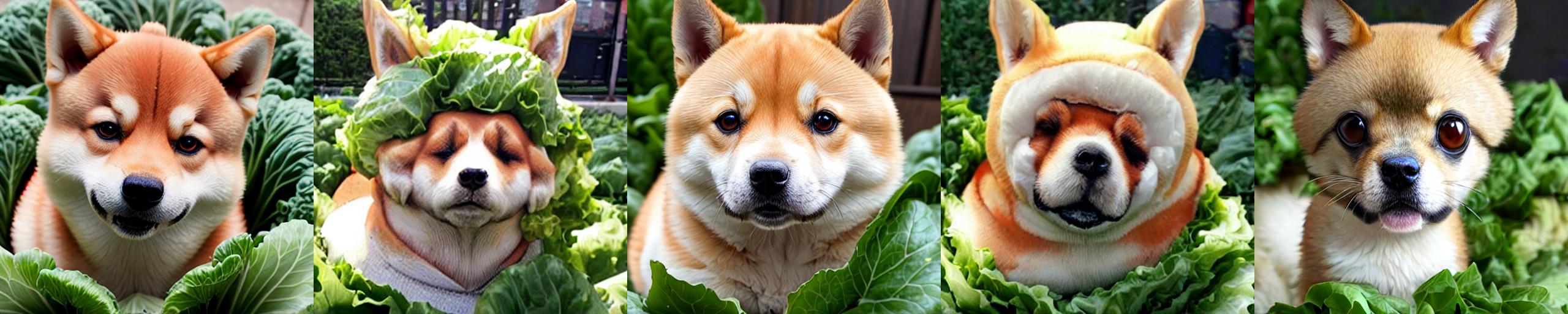}}
    \end{minipage}\\
    
    \begin{minipage}{0.1\textwidth}
    a giant cat carrying a small castle on its back, painting
    \end{minipage}\hspace{10pt}
    \begin{minipage}{0.64\textwidth}
	\subfloat{\includegraphics[width=\textwidth]{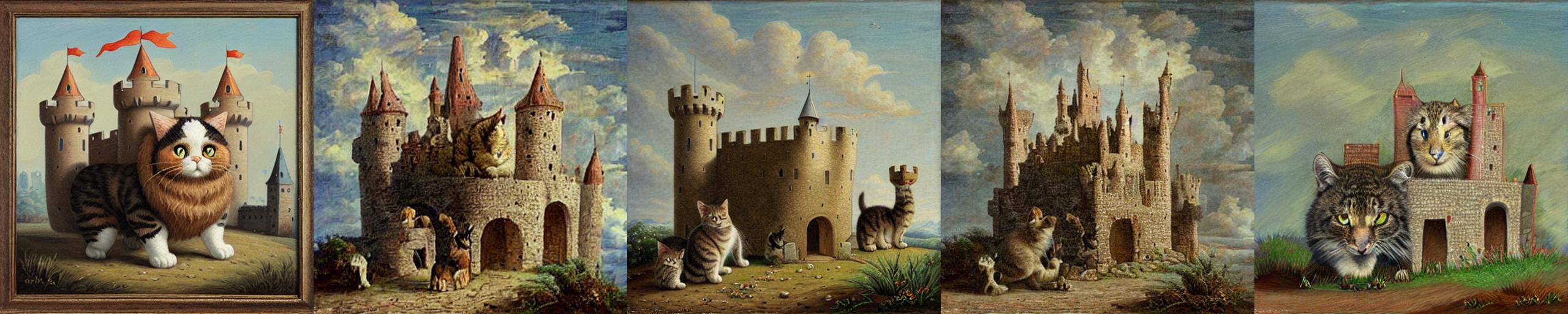}}
    \end{minipage}\\
    
    \begin{minipage}{0.1\textwidth}
    a classical guitar player at the concert
    \end{minipage}\hspace{10pt}
    \begin{minipage}{0.64\textwidth}
	\subfloat{\includegraphics[width=\textwidth]{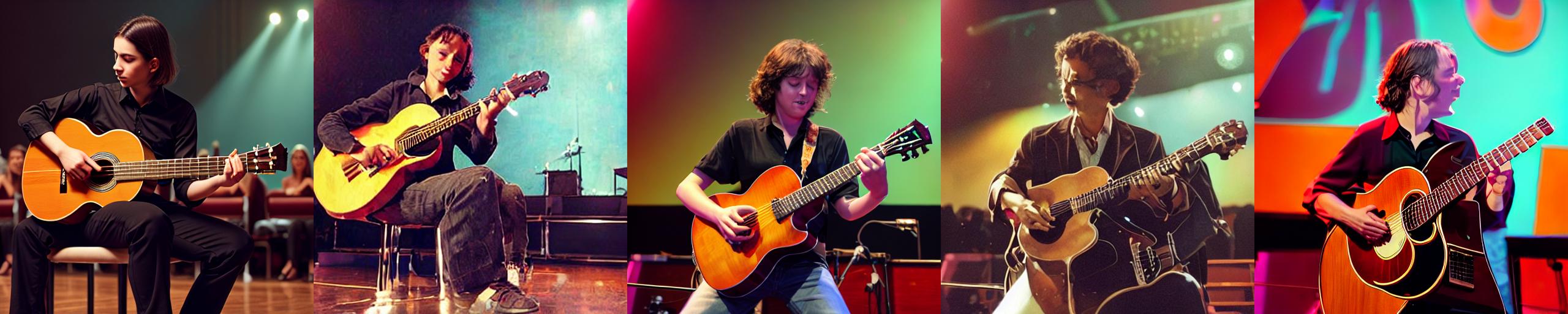}}
    \end{minipage}\\
    
    \begin{minipage}{0.1\textwidth}
    two tango dancers are dancing on the dim stage
    \end{minipage}\hspace{10pt}
    \begin{minipage}{0.64\textwidth}
	\subfloat{\includegraphics[width=\textwidth]{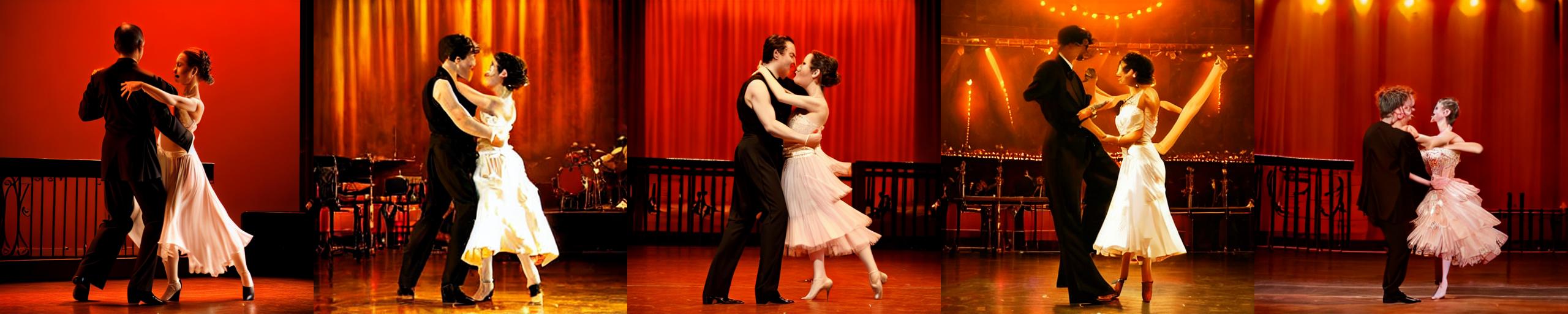}}
    \end{minipage}\\
\caption{Visual comparison with BK-SDM, implemented using RealisticVision-v4.0(RV40) as teacher, in DPM++ 25 steps.}
\label{fig_visual_comp_bk}
\end{figure*}

The visual comparisons with KOALA\cite{koala} and BK-SDM\cite{bk_sdm} are shown in Fig.\ref{fig_visual_comp_koala} and Fig.\ref{fig_visual_comp_bk}, respectively.
We can observe that our method achieved better visual results on different prompts compared with other methods.
Notably, it is also observed that our 55\% compressed SDXL-Base-1.0 and 33\% compressed RealisticVision-v4.0 can achieve almost the same visual qualities of the original models.

\end{document}